\documentclass[10pt,twocolumn,letterpaper]{article}

\usepackage{cvpr}              
\definecolor{cvprblue}{rgb}{0.21,0.49,0.74}
\usepackage[breaklinks,colorlinks,allcolors=cvprblue]{hyperref}
\usepackage[export]{adjustbox}
\usepackage{amsmath}
\usepackage{newtxtext,newtxmath}
\usepackage{xspace}
\usepackage{amssymb}
\usepackage{multirow}
\usepackage{multicol}
\usepackage{graphicx,caption,subcaption}
\usepackage[accsupp]{axessibility}
\usepackage{hyperref}
\usepackage{pifont}
\newcommand{\xmark}{\ding{55}}%
\newcommand{\cmark}{\ding{51}}
\newcommand{\OurBenchmark}
{\textsc{DrawingVQA}\xspace}

\def\paperID{22087} 
\def\confName{CVPR}
\def\confYear{2026}

\definecolor{cvprblue}{rgb}{0.21,0.49,0.74}

\usepackage{listings}
\usepackage{xcolor}

\lstdefinestyle{clean}{
    backgroundcolor=\color{gray!5},
    basicstyle=\ttfamily\small,
    numbers=none,
    numberstyle=\tiny\color{gray},
    keywordstyle=\color{blue},
    commentstyle=\color{green!50!black},
    stringstyle=\color{orange},
    breaklines=true,
    frame=single,
    rulecolor=\color{gray!50}
}\usepackage{multirow}
\usepackage{multicol}
\usepackage{graphicx,caption,subcaption}
\usepackage{hyperref}
\usepackage[most]{tcolorbox}
\usepackage{pifont}
\usepackage{CJKutf8}

\def\paperID{XXXXX} 
\def\confName{CVPR}
\def\confYear{2026}

\title{\OurBenchmark: A Real-World Benchmark for Multi-Depth Visual–Textual Reasoning on Construction Drawings}

\author{Yoonhwa Jung\thanks{Equal contribution $^\dagger$Work done during graduate studies at the University of Illinois Urbana-Champaign.}\\
\normalsize Louisiana State University\\
{\tt\small yoonhwa.jung@lsu.edu}
\and
Junryu Fu\footnotemark[1]$^\dagger$\\
\normalsize Independent Researcher\\
{\tt\small junryu.fu@gmail.com}
\and
Mani Golparvar-Fard\\
\normalsize University of Illinois Urbana-Champaign\\
{\tt\small mgolpar@illinois.edu}
}

\begin{document}
\maketitle

\begin{abstract}
We introduce \OurBenchmark, the first benchmark designed to evaluate multimodal large language models (MLLMs) on real-world construction drawings—a core media in architecture, civil, and many other engineering practices. Unlike natural images or schematic floor plans, construction drawings fuse abstract geometry, symbolic notation, tabular data, annotations, and domain-specific text, forming a uniquely complex visual–textual domain core to engineering workflows. \OurBenchmark bridges this gap with 33 “Issued for Construction” drawings and 92 expertly curated question–answer pairs, spanning three reasoning depths: perceptual understanding, contextual interpretation, and domain-expert reasoning. To evaluate model capabilities, we present a dual categorization framework to jointly analyze performance across seven construction-engineering and four MLLM capability dimensions-- the first to explicitly map engineering workflows to AI reasoning competencies. Evaluations of state-of-the-art MLLMs reveal a substantial gap between model and expert performance, particularly at higher reasoning depths. This benchmark lays a foundation for domain-specialized multimodal reasoning to allow for advancement on integration of AI-driven understanding and real-world engineering workflows. \OurBenchmark is available here: \url{https://joonv2.github.io/DrawingVQA/}

\end{abstract}

\vspace{-10pt}\section{Introduction}
\label{sec:intro}

\begin{figure*}[ht]
    \centering
    \includegraphics[width=0.96\textwidth]{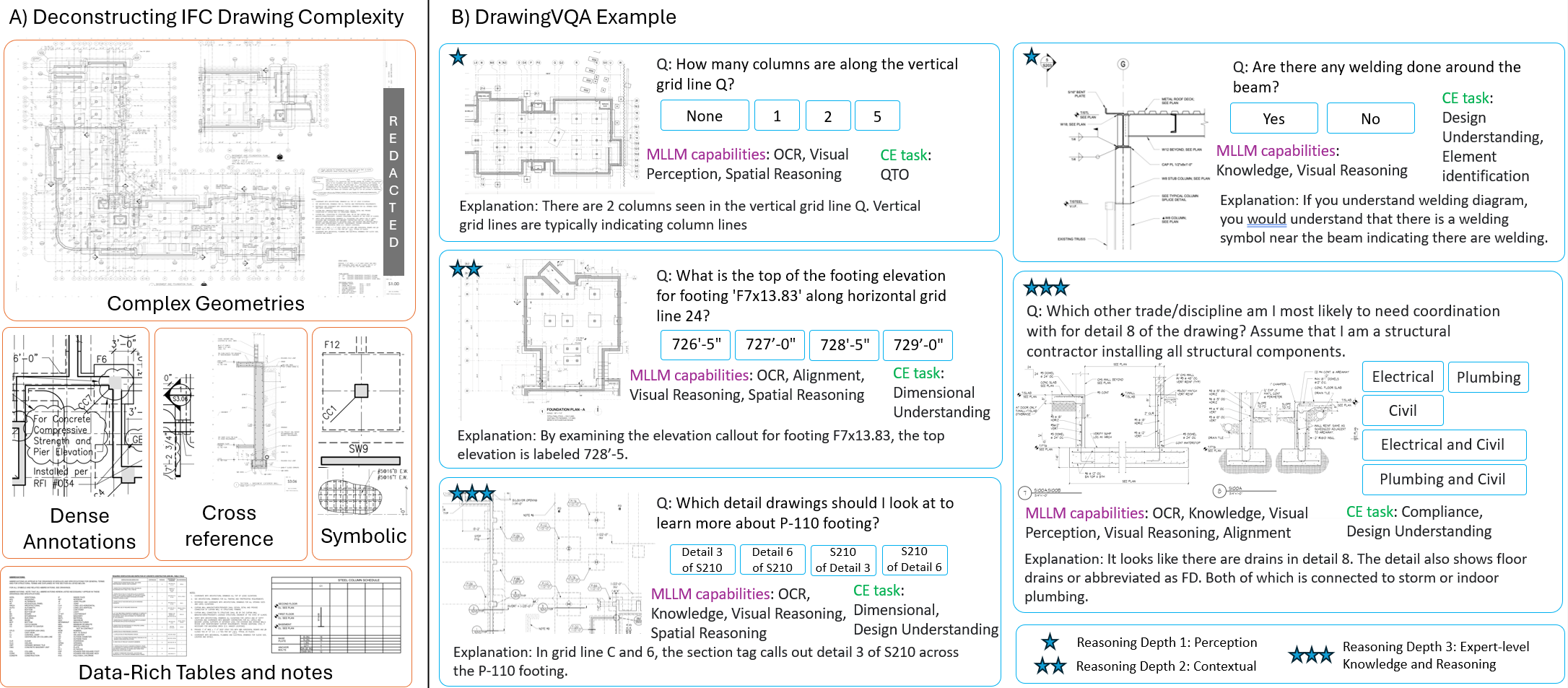}
    \caption{
    \textbf{The Multimodal Challenges of IFC Drawings and \OurBenchmark's Engineering-Level VQA.}
    \textbf{(A) Deconstructing IFC Drawing Complexity:} Real-world ``Issued for Construction" (IFC) drawings are information-dense artifacts that present unique multimodal challenges (See Section~\ref{sec:drawing}).
    \textbf{(B) \OurBenchmark VQA Examples:} We present challenging question-answer pairs from \OurBenchmark\ that demand industry-professional engineering reasoning, categorized by our three Reasoning Depths (indicated by stars: $\star$ Perception, $\star\star$ Contextual, $\star\star\star$ Expert-level Knowledge and Reasoning). Each example maps to specific \textbf{MLLM capabilities} (e.g., OCR, Visual Perception, Knowledge, Reasoning) and \textbf{Construction Engineering (CE) tasks} (e.g., QTO, Design Understanding, Element Identification, Compliance). These tasks extend far beyond academic benchmarks (See Section~\ref{sec:dual}).
}
\label{fig:overview}\vspace{-5pt}
\end{figure*}

The rapid advancement of Multimodal Large Language Models (MLLMs) has revealed a significant milestone toward artificial general intelligence (AGI), demonstrating remarkable capabilities that often match or even surpass human performance on perception, reasoning, and cross-modal understanding \cite{achiam2023gpt}. Recent multimodal benchmarks have shown that frontier models can approach or even surpass human performance in many generalist \cite{MMBench,mm-vet,A-okvqa,li2024seed}, academic \cite{mmmu,ScienceQA,zhang2023m3exam,rbench}, or professional fields such as finance \cite{gan2025mme,finmme}, medical \cite{VQA-Med,Omnimedvqa,liu2025gemex}, and biology \cite{microvqa,lozano2024micro}. However, deep engineering disciplines, particularly civil and construction engineering, remain largely underexplored.

Civil and construction engineering are among the most established yet least digitized engineering domains, where professional reasoning heavily depends on construction drawings—dense, multi-layered visual artifacts that combine geometric entities, annotations, tables, notes, engineering symbols, and cross-sheet references to communicate design intent and engineering problems. Existing research in the domain has mostly addressed low-level perceptual tasks, such as object detection or segmentation of simple floor plans \cite{floorplancad,r3d,RFP}. The few existing QA datasets for engineering disciplines tend to focus on academic knowledge, such as ``FE exam" style questions \cite{CMLLM,wu2025cequest}. This raises a critical question: \textit{Do MLLMs that excel at academic exams possess the practical reasoning skills of industry professionals?} And more broadly, \textit{To what extent do high scores on general benchmarks translate to competence in a complex, high-stakes domain like construction?}

To address this gap, we introduce \OurBenchmark, the first benchmark designed to evaluate MLLMs on real-world construction drawings. Unlike previous datasets, DrawingVQA uses ``Issued for Construction” (IFC)–grade drawings, reflecting the standards, complexity, and semantics encountered by practicing civil, structural, and construction engineers. The dataset challenging multi-image and interleaved text-image VQA questions comprises 33 drawings paired with 92 expertly curated question–answer pairs, reflecting the kinds of reasoning tasks engineers perform in real-world workflows. Each question is classified into three reasoning levels—easy (perceptual recognition), moderate (contextual interpretation), and difficult (expert reasoning)—capturing how humans progressively reason from perception to professional judgment. 

Beyond conventional accuracy metrics, DrawingVQA introduces a dual-categorization framework to analyze model performance jointly across seven construction-engineering dimensions (e.g., Quantity Take-Off, Design Intent, Code and Specification Compliance) and four core MLLM capability dimensions (Visual Perception, Knowledge, Reasoning, and OCR/Text Understanding). This multi-axis evaluation is the first to explicitly map engineering workflows to AI reasoning competencies, revealing where current models succeed, where they fail, and why.

It is the first benchmarking to evaluate open-source and proprietary MLLMs on IFC drawings with engineering professional levels. We summarize our contributions as follows:

\begin{itemize}
    \item We introduce \OurBenchmark, the first VQA benchmark built on complex, real-world ``Issued for Construction" (IFC) structural drawings, with carefully curated QA pairs reflecting professional reasoning depth.
    \item We propose a dual categorization framework that explicitly links professional engineering workflows to core AI reasoning competencies, enabling multidimensional analysis of model strengths and failures.
    \item We provide a comprehensive analysis of MLLMs, highlighting a critical gap in high-level engineering reasoning and setting a new, challenging baseline for future research.
\end{itemize}

\section{Related Work}
\label{sec:relatedwork}

\textbf{MLLM reasoning benchmarks}
The evaluation of MLLMs has evolved from general-perception VQA on natural images (e.g., VQA \cite{antol2015vqa}, GQA \cite{hudson2019gqa}, A-OKVQA \cite{A-okvqa}) to benchmarks testing deep, domain-specific expertise. This \textit{expert shift} requires deep disciplinary understanding, by testing college-level academic knowledge (e.g., MMMU \cite{mmmu}, SceMQA \cite{scemqa}) and targeting research-level reasoning in professional fields such as medicine \cite{VQA-Med,Omnimedvqa,PMC-VQA} and biology \cite{microvqa,lozano2024micro}. While this trend toward deep reasoning is critical, it has not yet addressed the unique, practical reasoning required for engineering, where visual and textual information are deeply intertwined through symbolic diagrams, geometric and spatial relationships, and domain-specific conventions.

\noindent\textbf{MLLM in Engineering.}
Existing MLLM benchmarks related to engineering predominantly evaluate academic or declarative knowledge, not practical skills. For example, the Engineering category of MMMU \cite{mmmu} consists of college-level exam-style questions using natural images, tables, or schematic diagrams. Similarly, R-Bench \cite{rbench} features graduate-level exam-style questions in engineering disciplines. ScienceQA \cite{ScienceQA} focuses on scientific experiment questions on natural and general images (i.e., photos) without domain-level visual grounding. As we argue, these datasets, while valuable, do not test comprehensive multimodal reasoning—such as cross-and multi-view spatial understanding, detail referencing, or interpreting technical annotations—skills essential for professional engineering tasks.

Closer to our domain, other benchmarks have partially addressed design-phase documents. MM-Vet~\cite{mm-vet} features VQA on simplified floor plans, analyzing OCR, spatial awareness, and math capabilities of MLLMs. DesignQA~\cite{designqa} performs VQA over mechanical CAD drawings along with additional PDF documents, while DrafterBench~\cite{drafterbench} introduces LLM-based editing on vectorized CAD data. While valuable, these datasets focus on schematic, low-density design representations and fall short of evaluating the multimodal reasoning required for Issued for Construction (IFC) drawings—documents that integrate geometric complexity, symbolic notation, tabular data, annotations, and discipline-specific text.

\noindent\textbf{Construction domain benchmarks}
Within the AEC domain, research has historically bypassed high-level reasoning on IFC drawings, focusing instead on other data types. A significant body of work targets low-level perceptual tasks on schematic floor plans, such as object detection, segmentation, 3D reconstruction  \cite{floorplancad,r3d,RFP,ArchCAD-400K} or image captioning \cite{vsd,chen2025}. Other benchmarks address text-only question answering based on licensing exams or professional guidelines \cite{CMLLM,wu2025cequest}. Such approaches completely bypass the multimodal nature of engineering communication. Consequently, there remains no benchmark that captures how engineers reason across visual, textual, and spatial modalities in construction professional workflows. \OurBenchmark is the first to address this gap with an expert-level VQA dataset grounded in real-world IFC drawings. Section~\ref{sec:compwithothers} shows how \OurBenchmark differ from other standard MLLM benchmarks in engineering disciplines.


\section{\OurBenchmark}

\subsection{Overview of \OurBenchmark}
The primary goal of \OurBenchmark is to address the significant gap between existing MLLM benchmarks and the practical, real-world visual reasoning tasks performed by engineering professionals. Current benchmarks often focus on academic knowledge or simplified schematics, failing to capture the complexity of industry-standard documents. \OurBenchmark is designed to identify and measure these performance gaps in frontier models, providing a crucial tool to drive industry adoption.

To achieve this, our benchmark is built on two core principles:
\begin{itemize}
    \item \textbf{Authentic Data Source:} We use real-world, ``Issued for Construction" (IFC) structural drawings, which are the dense, multi-modal, and legally-binding documents used in the field.
    \item \textbf{Practical Questioning:} The question-answer (QA) pairs are crafted by three domain experts to resemble the day-to-day queries and reasoning workflows professionals perform, not just surface-level academic questions.
\end{itemize}

The dataset spans three distinct reasoning depths: (1) simple perceptual understanding, (2) contextual interpretation, and (3) deep domain-expert reasoning. To analyze model performance with high granularity, we introduce a novel dual categorization framework that jointly maps each QA pair across seven construction-engineering dimensions and four foundational MLLM capability dimensions. This is the first framework, to our knowledge, to explicitly map engineering workflows to core AI reasoning competencies, offering diagnostic insights into why and where models fail in real-world construction engineering contexts.

\vspace{-4pt}
\begin{table}[h!]
\centering
\caption{\OurBenchmark statistics}
\label{tab:stats}
\vspace{-2pt}
\begin{adjustbox}{width=0.45\textwidth,keepaspectratio}
\begin{tabular}{lc}
\toprule
Statistics & Value \\
\midrule
Total questions & 92 \\
\midrule
Image (artifacts) types & 5 \\
Total images & 58 \\
Unique drawing & 33 \\
\midrule
Multiple-choice questions & 90 \\
Open questions & 2  \\
Questions with an explanation & 100\% \\
\midrule
Image in Question  & 97.78\% \\
 \hspace{1em} * Image at the beginning & 24.72\% \\
 \hspace{1em} * Image in the middle & 34.96\% \\
 \hspace{1em} * Image at the end & 39.33\% \\
Image in Options & 2.22\% \\
Example with Multiple Images & 14.44\% \\
\midrule
Reasoning Level (Perceptual:Contextual:Expert-Reasoning) & 36.96\%:33.70\%:29.35\% \\
\midrule
Questions Mapped to MLLM Capabilities & \\
\hspace{1em} * Visual Perception &  17.73\% \\
\hspace{1em} * Reasoning (Total) & 29.06\%   \\
\hspace{3em} * Visual reasoning &  39.3\% \\
\hspace{3em} * Alignment and grounding & 32.1\%   \\
\hspace{3em} * Spatial reasoning &  28.6\% \\
\hspace{1em} * Knowledge & 15.76\% \\
\hspace{1em} * OCR (Text Understanding) & 37.44\% \\
Questions Mapped to Construction Eng. Dimensions &  \\
\hspace{1em} * General administration & 8.94\% \\
\hspace{1em} * Domain-specific element identification & 20.67\% \\
\hspace{1em} * Domain-specific language understanding & 13.41\% \\
\hspace{1em} * Design semantic understanding & 29.05\% \\
\hspace{1em} * Dimensional relationship reasoning & 17.32\% \\
\hspace{1em} * Quantity takeoff & 6.70\% \\
\hspace{1em} * Compliance check & 3.91\% \\
\midrule
Average question length & 14 words\\
Average time to create 1 question & 38 mins \\
Average time to quality check 1 question & 24 mins \\

\bottomrule
\end{tabular}
\end{adjustbox}
\end{table}

\vspace{-5pt}
\begin{figure}[h]
	\centering
	\includegraphics[width=0.45\textwidth]{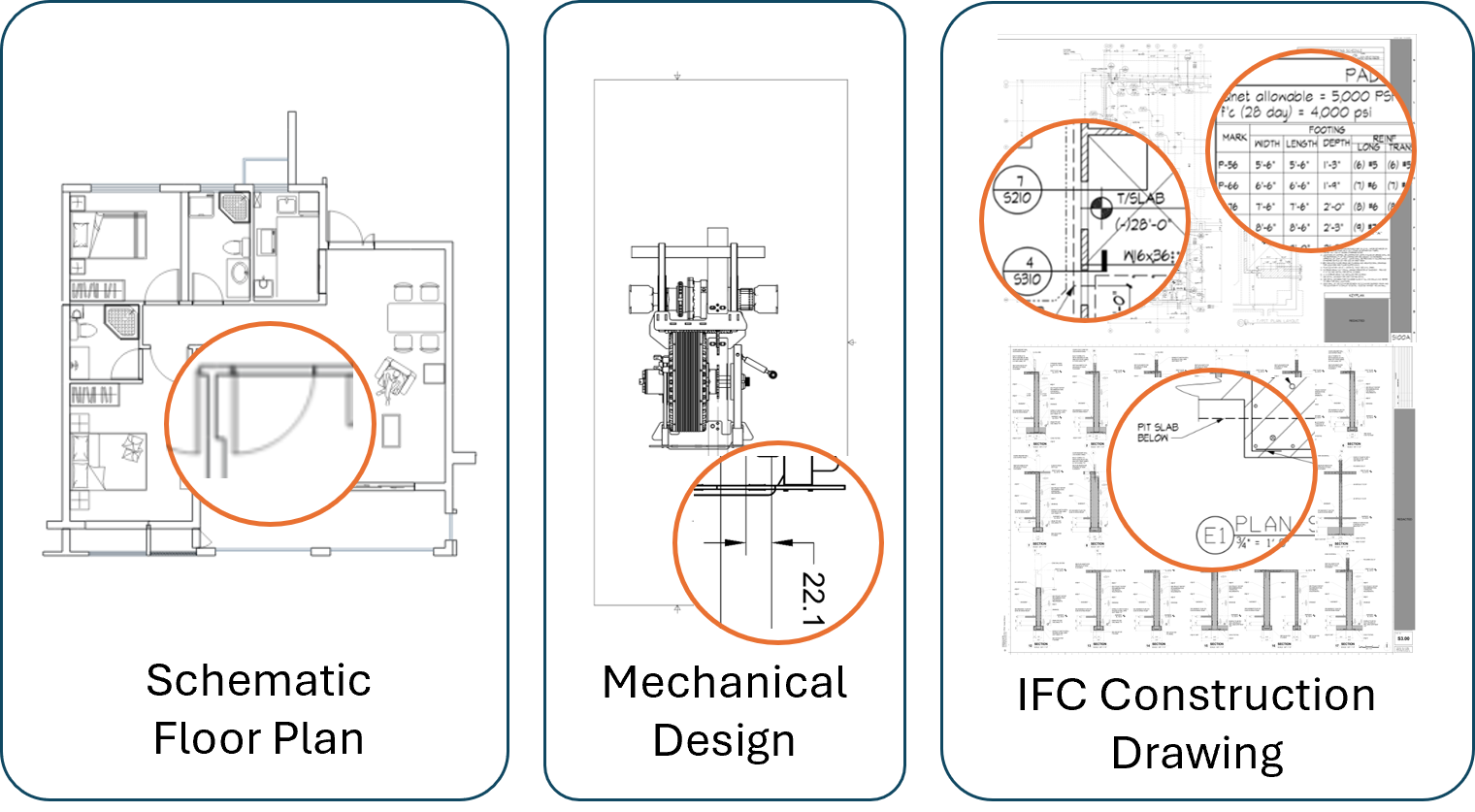}
    \caption{Comparison of drawing complexity. (a) A low-density, schematic architectural floor plan from \cite{floorplancad}. (b) A mechanical design from \cite{designqa}. (c) A high-density, ``Issued for Construction" (IFC) level drawing from our \OurBenchmark corpus.}
	\label{fig:drawing-compare}\vspace{-5pt}
\end{figure}

\subsection{Construction Drawings (Issued for Construction)}\label{sec:drawing}

The visual foundation of \OurBenchmark consists of 33 structural drawing sets from six real-world educational building projects. Our corpus is composed of high-entropy ``Issued for Construction'' (IFC) documents. It is crucial to differentiate the visual and semantic complexity of IFC drawings from the schematic architectural floor plans or mechanical design used in other datasets (see Figure \ref{fig:drawing-compare}). Compared to other 2D drawings, an IFC drawing spans the entire project lifecycle as ``source of truth" that integrates precise geometry, engineering annotations, material specifications, revision histories, and regulatory compliance notes—all of which form a dense visual–textual interface that professionals interpret daily.

In details, IFC drawings represent a unique multimodal challenge precisely because they are dense, information-rich artifacts that layer:
\begin{itemize}
    \item \textbf{Complex Geometries:} Precise 2D representations (plans, sections, elevations) of structural elements (Figure~\ref{fig:overview}).
    \item \textbf{Symbolic Language:} Symbols and domain-specific vocabulary (e.g., weld symbols, material hatches, discipline specific language and acronyms)
    \item \textbf{Dense Annotations:} Textual and symbolic callouts, dimensions, and annotations with arrows that require high-fidelity OCR.
    \item \textbf{Data-Rich Tables and Notes:} Complex tabular schedules for columns, beams, footings, and revisions and general notes.
    \item \textbf{Cross-References:} Pointers that require a model to align information and understand different dimensional views across different details and multiple sheets with callouts.
\end{itemize}


\subsection{Data Curation Process}


\noindent\textbf{Data Collection}
We curated our dataset from a collection of 33 structural drawing sets from 6 campus-building projects, ensuring each document meets professional Issued for Construction (IFC) standards. The visual artifacts include plans, sections, details, annotations, symbols, and general notes covering diverse structural components (e.g., slabs, columns, beams, and foundations). All drawings used in the dataset conform to a common industry standard for publishing IFC drawings called the U.S. National CAD Standard (NCS): core elements such as symbols, callouts, sheet IDs, and table formats remain consistent across all construction and facility design projects in the United States. The NCS standard ensures consistency with professional drafting conventions and serves as the foundational rationale for interpreting symbols, annotations, and callouts within the dataset, aligning our benchmark with established industry documentation practices. All QA pairs were generated by a team of three domain experts, including licensed civil engineers and construction professionals with 2 to 10+ years of field experience. The process was designed to emulate the authentic, information-seeking behaviors of construction professionals.

\noindent\textbf{Reasoning Depth} Experts were first instructed to formulate questions spanning three progressive reasoning depths. The first level, Perceptual Understanding, includes simple, direct questions requiring visual search or basic OCR (e.g., \textit{``What is the sheet title?"}). The second, Contextual Interpretation, requires models to connect and reason about multiple pieces of information on a single sheet (e.g., \textit{``What is the top of the footing elevation for footing `F7x13.83' along horizontal grid line 24?"}). The final and most complex level, Domain-Expert Reasoning, demands compositional reasoning, implicit domain knowledge, or aligning information across different drawings (e.g., \textit{``What type of vertical reinforcements would I need for the walls shown? Use the other drawing if needed."}). All answers were required to be concise, factual, and directly verifiable from the provided drawing set.

\noindent\textbf{VQA Format} Each question was formulated as either a multiple-choice or a short open-answer question. For the multiple-choice questions, our experts manually created all distractors to avoid low-quality, easily guessable, or LLM-generated options. Distractors often included realistic but subtly incorrect alternatives such as picking up nearby texts on the drawings that are actually irrelevant or challenging options such as ``None of the above” or ``All of the above”
This expert-driven design ensures the distractors reflect authentic misinterpretations of IFC drawings, serving as a rigorous proxy for professional ambiguity. Finally, comparative testing between the MCQ and open-ended formats confirms that the closed-ended structure does not artificially reduce task difficulty (further details are provided in the Supplementary).
This multi-stage, expert-driven process produced 92 question-answer pairs grounded in real construction drawings—capturing not only the visual and textual complexity of engineering documents but also the depth of reasoning required in professional practice.

\noindent\textbf{Quality Control and Validation}
To ensure dataset integrity, we employed a rigorous, multi-pass validation protocol. First, each QA pair generated by one expert was independently reviewed by a second expert for correctness, ambiguity, and practical relevance. Any disagreements were escalated to a senior engineer for final adjudication. Second, we explicitly filtered out any questions that could be answered without the image or were unanswerable with the provided context. Finally, after validation, every QA pair was annotated according to our dual categorization framework, which jointly captures (a) the construction-engineering domain aspect of the inquiry and (b) the MLLM capability dimension required to solve the question. These annotations underwent the same multi-pass review cycle to maintain labeling reliability and conceptual clarity.

\subsection{Dual Categorization Framework}\label{sec:dual}
The core of \OurBenchmark evaluation comes from our dual categorization framework. Each QA pair is tagged along two orthogonal axes, allowing for a multifaceted analysis of model capabilities.

\noindent\textbf{1. MLLM Capability Dimension (The ``AI Task")}
This axis categorizes the foundational AI competency required to answer the question. It is divided into four primary dimensions: \textbf{Visual Perception} (simple identification, attribute recognition, and counting); \textbf{Reasoning} (higher-order cognition, which is further subdivided into \textit{Visual Reasoning}, \textit{Alignment \& Grounding}, and \textit{Spatial Reasoning}); \textbf{Knowledge} (accessing implicit, domain-specific information not explicitly written on the sheet); and \textbf{OCR} (text recognition and understanding, inside \textit{Tabular} for schedules and \textit{Textual} for callouts and notes). When a task requires only reading text, its MLLM capability dimension is OCR alone. If the question necessitates aligning that text from tabular with geometric entities or layouts, it is explicitly mapped to both OCR and Reasoning category (Visual, Spatial, Alignment).

\noindent\textbf{2. Construction-Engineering Dimension (The ``Practical Task")}
This axis categorizes the real-world engineering workflow the question belongs to. The seven dimensions are: \textbf{General / Administrative} (reading metadata, tables, and general notes); \textbf{Element Identification} (interpreting structural symbols); \textbf{Language Understanding} (interpreting domain-specific languages); \textbf{Dimensional Understanding} (cross-referencing on multi-view drawing parts and relationships); \textbf{Design Semantic Understanding} (inferring drawing type or design intent); \textbf{Quantity Take-Off (QTO)} (tasks related to counting elements); and \textbf{Compliance} (verifying components against a code or specification).

We provide a detailed breakdown of the dataset statistics and the distribution of these categories in Table \ref{tab:stats}. This dual framework allows us to move beyond a single accuracy score and precisely diagnose why a model fails for engineering workflows, providing a clear roadmap for future improvements.

\begin{figure}
    \centering
    \includegraphics[width=0.98\linewidth]{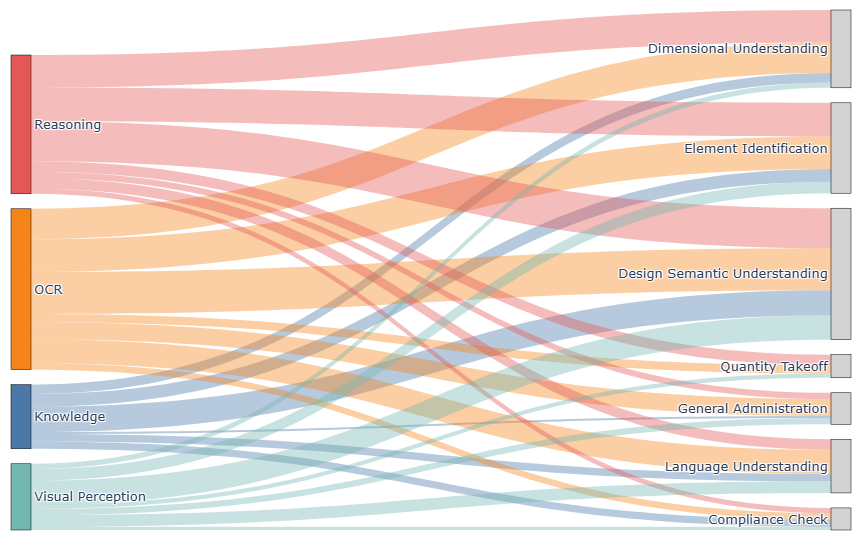}
    \caption{Visualization of the dual categorization mapping MLLM Capabilities (left) to their corresponding Construction-Engineering Dimensions (right) in \OurBenchmark. The flow diagram highlights how complex engineering tasks, such as ``Dimensional Understanding", draw upon multiple MLLM Reasoning and OCR capabilities.}
    \label{fig:dual}\vspace{-5pt}
\end{figure}


\begin{figure*}[h]
    \centering
    {\includegraphics[width=0.25\linewidth,valign=c]{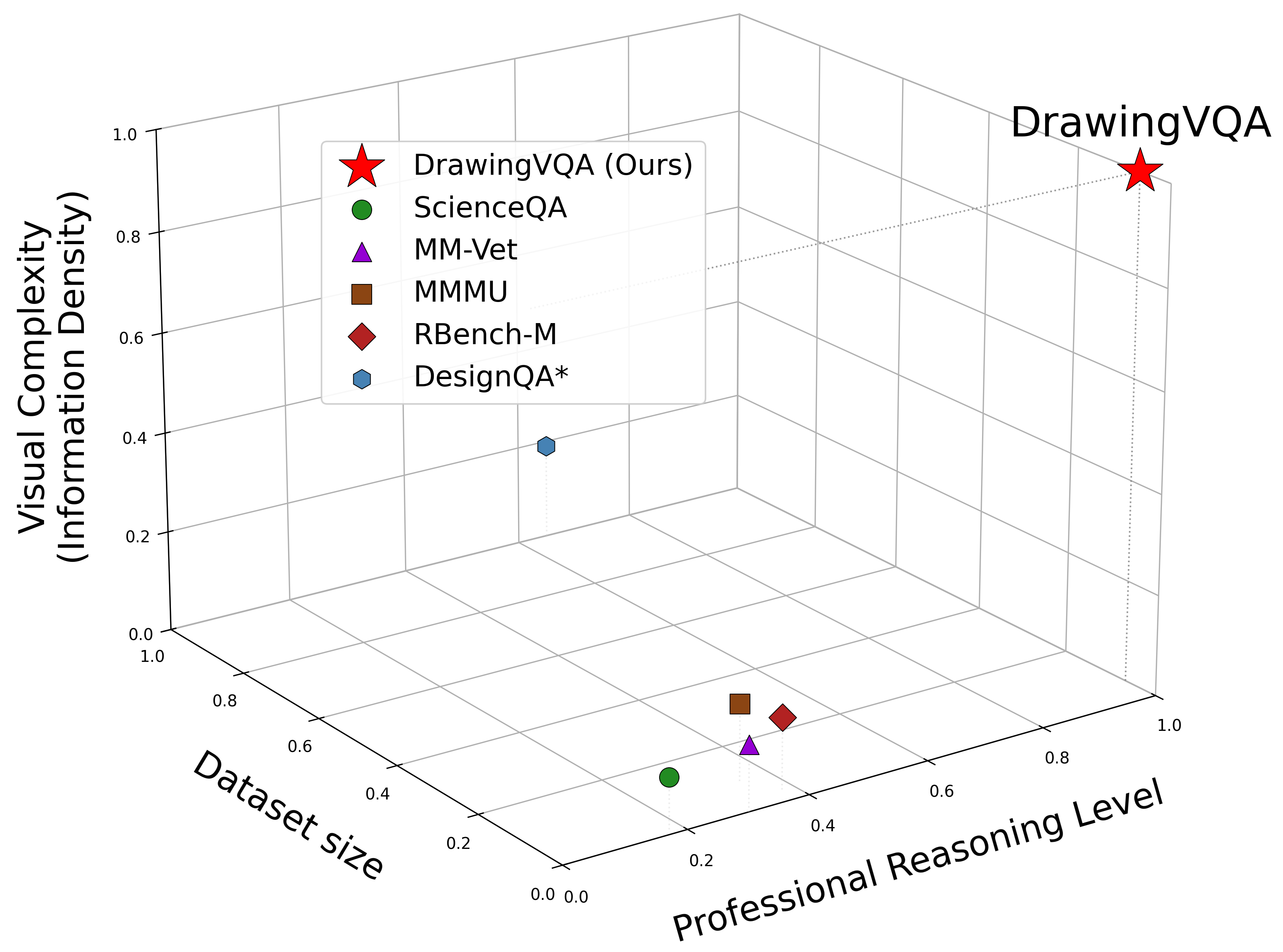}}
    {\adjustbox{width=0.73\textwidth,keepaspectratio,valign=c}{\begin{tabular}{@{}l l l l c l c r@{}}
            \toprule
            Benchmark & Level & Visual source & Option creation & Q.diversity & Answer & Expln.  &Size \\
            \midrule
            \textbf{\OurBenchmark (Ours)} & \textbf{Eng. Profl.} & \textbf{Real-world IFC drawings} & \textbf{Annotation} & 7 & \textbf{MC + Open} & \textbf{\cmark} & \textbf{92} \\
            ScienceQA (subset) \cite{ScienceQA} & Pre-college & K–12 curricula & Textbook & 1 & MC & \xmark & 24 \\
            MM-Vet (subset) \cite{mm-vet} & Undergraduate & Floor plan & Annotation & 1 & Open & \xmark & 4 \\
            MMMU (subset) \cite{mmmu} & Undergraduate & Textbooks, WebQA & Textbook & 2 & MC + Open & $\triangle$ & 135 \\
            RBench-M (subset) \cite{rbench} & Graduate & Exams & Exams & 1 & MC & \xmark & 54 \\
            DesignQA$^{*}$ \cite{designqa} & Eng. Profl. & Mechanical drawings & Annotation & 3 & Open & \xmark & 1451 \\
            \bottomrule
            \end{tabular}}}
        \caption{Comparison of \OurBenchmark with other VQA benchmarks, which contain any of Architecture, Civil, Structural, or Construction Engineering disciplines, or drawing-related questions. Q. means Question, Expln. means Explanations, Profl. means Professional, Eng. means Engineering, MC means multiple-choice, and Open means open-ended. DesignQA is not a pure VQA dataset, with document-reference dependence. Compared to others, \OurBenchmark has the highest visual complexity (also featured in Figure~\ref{fig:drawing-compare}) and the highest professional reasoning level in engineering practices.}
        \label{fig:comparison}\vspace{-5pt}
    \end{figure*}

\subsection{Comparisons with Existing Benchmarks}\label{sec:compwithothers}
\OurBenchmark occupies a unique position at the intersection of multimodal reasoning and engineering document packed with domain specific knowledge. As summarized in Figure~\ref{fig:comparison}, existing benchmarks that include AEC content address only narrow, academic aspects of the discipline. MMMU~\cite{mmmu} contains a small subset of 135 civil and structural engineering-related questions framed as textbook problems. Similarly, RBench-M \cite{rbench} includes 54 questions from structural engineering textbook exams. ScienceQA~\cite{ScienceQA} offers 24 relevant engineering practice questions based on natural images, and MM-Vet~\cite{mm-vet} includes only four schematic floor plans for basic spatial reasoning. These datasets evaluate declarative knowledge rather than assessing context specific multimodal reasoning required in real construction workflows. 

One notable VQA benchark, DesignQA~\cite{designqa}, focuses on the mechanical and industrial engineering domain, which possess parallels with DrawingVQA on 2D drawing and compliance pdf document on compliance check. However, the QA images utilized were cleaned of any confusing annotations, and other textual or tabular, and symbolic visuals that hinder perception performance. Additionally, the majority of their QA dataset relies heavily on simple, text-based measurement checks against PDF documents. These ``hindrances" were kept true to its color in the \OurBenchmark, just as anyone sees everyday at real construction projects, requiring authentic, real-world engineering knowledge rather than basic text extraction.

\OurBenchmark is the first dataset built entirely on real IFC-level structural drawings, reflecting how professionals interpret, cross-reference, and reason across sheets in daily practice. To avoid artificially inflating model performance with repetitive, low-level queries (e.g., repeatedly extracting sheet IDs), we curated a \underline{high-entropy dataset} that maximizes diagnostic coverage. Each question targets a specific failure mode and distinct reasoning pathway, ensuring the task demands the multi-step inferential workflows required at jobsites rather than reducing to simple visual grounding. Finally, by incorporating a dual categorization framework spanning both construction-engineering and MLLM capability dimensions, it performs fine-grained diagnosis of model reasoning gaps, establishing a foundation for advancing AI toward domain-grounded engineering intelligence.


\vspace{-3pt}
\section{Experiments}

\subsection{Baselines}
\textbf{MLLMs} To evaluate \OurBenchmark, we benchmarked a range of state-of-the-art and recent MLLMs. This includes leading proprietary models such as OpenAI's o3 and GPT-4o, Google's Gemini 2.5 Pro and 2.5 Flash, and Anthropic's Claude-4.5 Sonnet and open-source models, including 
LLaVA-OneVision-1.5-8B-Instruct \cite{OneVision-1.5}, LLaVA-v1.6-Mistral-7B \cite{liu2023improved}, Llama-3.2-11B-Vision-Instruct \cite{Llama32}, Qwen3-VL-8B-Instruct \cite{qwen3technicalreport}, 
InternVL3.5-30B-A3B\cite{wang2025internvl3_5},
and Phi-4-multimodal-instruct \cite{microsoft2025phi4}, which are known for their strong performance on general vision-language tasks. All models were tested using standard chain-of-thought prompting in a zero-shot setting and answer parsing, following the protocol established in MicroVQA \cite{microvqa}. All experiments are conducted with an NVIDIA A100 GPU.

\noindent\textbf{Human Benchmarking and Evaluation}
To benchmark MLLM performance against true domain expertise, we established a robust human baseline. We randomly selected a subset of \texttt{20} QA pairs from our test set, stratified across the three reasoning depths (4 Perceptual, 8 Contextual, 8 Expert). 52 Responses were collected from three groups representing increasing levels of domain expertise: (i) undergraduate civil engineering students (not in their freshman year, mainly junior and seniors), (ii) graduate civil engineering students and/or early-career professionals (less than three years of industry experience), and (iii) experienced professionals (three or more years of experience). This allows us to not only compare models to peak human performance but also to understand how model reasoning capabilities align with different stages of professional development. To maintain a fair experiment, all the QA answering were timed (20 minutes) to ensure all participants had equal amount of time to answer the same number of questions. In addition, we added Random Choice baseline for reference.

\noindent\textbf{Evaluation}
We report overall mean accuracy for both multiple-choice and open-ended questions. In addition, we provide fine-grained analysis across both MLLM capability dimensions (visual perception, reasoning, knowledge, OCR) and construction-engineering dimensions (general, element identification, language understanding, dimensional reasoning, semantic understanding, quantity takeoff, code/specification compliance). More details on the dimensions can be found in the Supplementary. We also evaluate model performance under no-image conditions and using randomized multiple-choice label prefixes (ABCD alphabetical order), reported in the Supplementary.

\begin{table*}[h!]
\centering
\caption{Model Performance with Reasoning, MLLM, and Construction Metrics. \textbf{MLLM Dimensions} (VP: Visual Perception, K: Knowledge, R: Reasoning, OCR: Optical Character Recognition); 
\textbf{Construction Eng. Domain Aspects} (Adm.: Administration, EI: Element Identification, LU: Domain Language Understanding, DR: Dimensional Recognition, SU: Design Semantic Understanding, QTO: Quantity Take Off, Comp.: Compliance). Underscored values indicate the highest score for each model within the MLLM Dimensions and CE Domain Aspects categories.}
\label{tab:my_model_comparison}
\begin{adjustbox}{width=1\textwidth,totalheight=0.9\textheight,keepaspectratio}
\begin{tabular}{lc|ccc|cccc|ccccccc}
\toprule
&& \multicolumn{3}{c|}{\textbf{Reasoning Depth}} & \multicolumn{4}{c|}{\textbf{MLLM Dimensions}} & \multicolumn{7}{c}{\textbf{Construction Eng. Domain Aspects}} \\
\midrule
\textbf{Model} & \textbf{Overall} & \textbf{R1} & \textbf{R2} & \textbf{R3} & \textbf{VP} & \textbf{K} & \textbf{R} & \textbf{OCR} & \textbf{Adm.} & \textbf{EI} & \textbf{LU} & \textbf{DR} & \textbf{SU} & \textbf{QTO} & \textbf{Comp} \\
\midrule


GPT-4o & 48.9 & 61.3 & 44.1 & 40.7 & 47.1 & \underline{59.4} & 44.3 & 48.7 & 56.3 & 32.4 & \underline{70.8} & 41.9 & 50.0 & 25.0 & 42.9 \\
o3 & 58.7 & 64.5 & 61.8 & 48.2 & 58.8 & \underline{65.6} & 54.1 & 59.2 & 62.5 & 56.8 & 54.2 & 48.4 & 61.5 & 25.0 & 42.9\\
\textbf{Gemini-2.5-pro} & \textbf{71.7} & 80.6 & 76.5 & 55.6 & 67.6 & \underline{78.1} & 67.2 & 72.4 & \underline{87.5} & 56.8 & 79.2 & 67.7 & 75.0 & 41.7 & 57.1\\
Gemini-2.5-flash & 66.3 & 77.4 & 73.5 & 44.4 & 64.7 & \underline{81.2} & 57.4 & 63.2 & \underline{87.5} & 51.4 & 79.2 & 51.6 & 67.3 & 33.3 & 71.4\\
Claude-4.5-Sonnet & 57.6 & 64.5 & 64.7 & 40.7 & 58.8 & \underline{59.4} & 57.9 & 54.1 & \underline{68.8} & 48.7 & 66.7 & 41.9 & 57.7 & 41.7 & 57.1\\
Claude-4.5-Haiku & 54.3 & 51.6  & 73.5  & 33.3  & 52.9  & \underline{65.6}  &  52.5 & 54.0  & \underline{62.5}  & 43.2 &  50.0 & 48.4 & 51.9  & 33.3   &42.9 \\
\midrule

LLaVA-OneVision-1.5-8B-Instruct \cite{OneVision-1.5} & 40.2 & 61.3 & 41.2 & 14.8 & 38.2 & \underline{46.9} & 31.1 & 39.5 & 43.8 & 24.3 & \underline{62.5} & 25.8 & 42.3 & 0.0 & 0.0\\
LLaVA-v1.6-Mistral-7B \cite{liu2023improved} & 28.3 & 25.8 & 29.4 & 29.6 & 29.4 & \underline{37.5} & 27.9 & 23.7 & 31.2 & 24.3 & 16.7 & \underline{32.3} & 30.8 & 25.0 & 14.3\\
Llama-3.2-11B-Vision-Instruct \cite{Llama32} & 39.1 & 45.2 & 47.1 & 22.2 & \underline{41.2} & 40.6 & 34.4 & 38.2 & \underline{50.0} & 29.7 & 45.8 & 32.3 & 42.3 & 8.3 & 28.6\\
Qwen3-VL-8B-Instruct \cite{qwen3technicalreport} & 53.3 & 51.6 & 61.8 & 44.4 & 50.0 & 53.1 & 45.9 & \underline{55.3} & \underline{81.2}& 40.5 & 66.7 & 45.2 & 46.2 & 33.3 & 42.9\\
InternVL3.5-30B-A3B\cite{wang2025internvl3_5} & 41.3 & 51.6 & 35.3 & 37.0 & 26.5 & \underline{43.8} &  42.6 & 42.1 & 50.0 &  40.5 & \underline{45.8} & 45.1 & 46.1 & 8.3 & 14.3 \\
Phi-4 \cite{microsoft2025phi4} & 40.2 & 41.9 & 47.1 & 29.6 & 29.4 & 40.6 & \underline{42.6} & 40.8 & 37.5 & 40.5 & \underline{45.8} & 35.5 & 40.4 & 25.0 & 28.6\\

\midrule

Random & 27.2 & 25.8 & 17.7 &  40.7 & 23.5 & \underline{37.5} &  27.9 & 29.0 & 37.5 &  18.9 & 29.2 & 29.0 &  23.0 & 16.6 & \underline{42.8} \\
Human & 68.4 & 75.6 & 72.6 & 59.9 & 75.8 & 81.4 & \underline{84.9} & 75.0 & 46.8 & 71.9 & 82.7 & 82.1 & \underline{99.2} & 83.3 & 66.2\\
\hspace{1em} * Undergraduate & 62.8 & 71.8 & 70.2 & 53.8 & 64.3 & 66.7 & \underline{74.9} & 53.0 & 43.6 & 49.7 & 70.7 & 75.4 & \underline{97.4} & 69.4 & 46.2\\
\hspace{1em} * Graduate \& Young Professionals & 78.2 & 85.4 & 76.6 & 68.8 & 73.9 & 81.9 & \underline{83.8} & 76.0 & 50.0 & 70.0 & 81.9 & 75.0 & \underline{100.0} & 83.7 & 62.5\\
\hspace{1em} * Professionals & \textbf{94.9} & 90.0 & 85.0 & 93.3 & 89.1 & 95.6 & \underline{96.0} & \underline{96.0} & 46.7 & 96.0 & 95.6 & 96.0 &\underline{100.0} & 96.6 & 90.0\\
\bottomrule
\end{tabular}
\end{adjustbox}
\end{table*}

\subsection{Main Results}
\textbf{Overall Performance: SOTA vs. Human Experts.}
A primary finding is the significant performance gap that still exists between most MLLMs and human-level expertise. Only one model, \texttt{Gemini-2.5-pro}, achieved an overall score of 71.7, surpassing the average human score (68.4) and the undergraduate student cohort (62.8). However, this still falls substantially short of the performance of experienced \textit{Professionals} (94.9), indicating that while SOTA models can match entry-level performance, they do not yet possess deep domain expertise. The next-best models, \texttt{Gemini-2.5-flash} (66.3) and \texttt{o3 (58.7)}, also demonstrate strong capabilities but, like most other models (e.g., \texttt{GPT-4o} at 48.9, \texttt{Claude-4.5-Sonnet} at 57.6), still struggle to match the baseline performance of an undergraduate student in civil engineering major.

\begin{figure}[htp]
    \centering
    \includegraphics[width=0.85\linewidth]{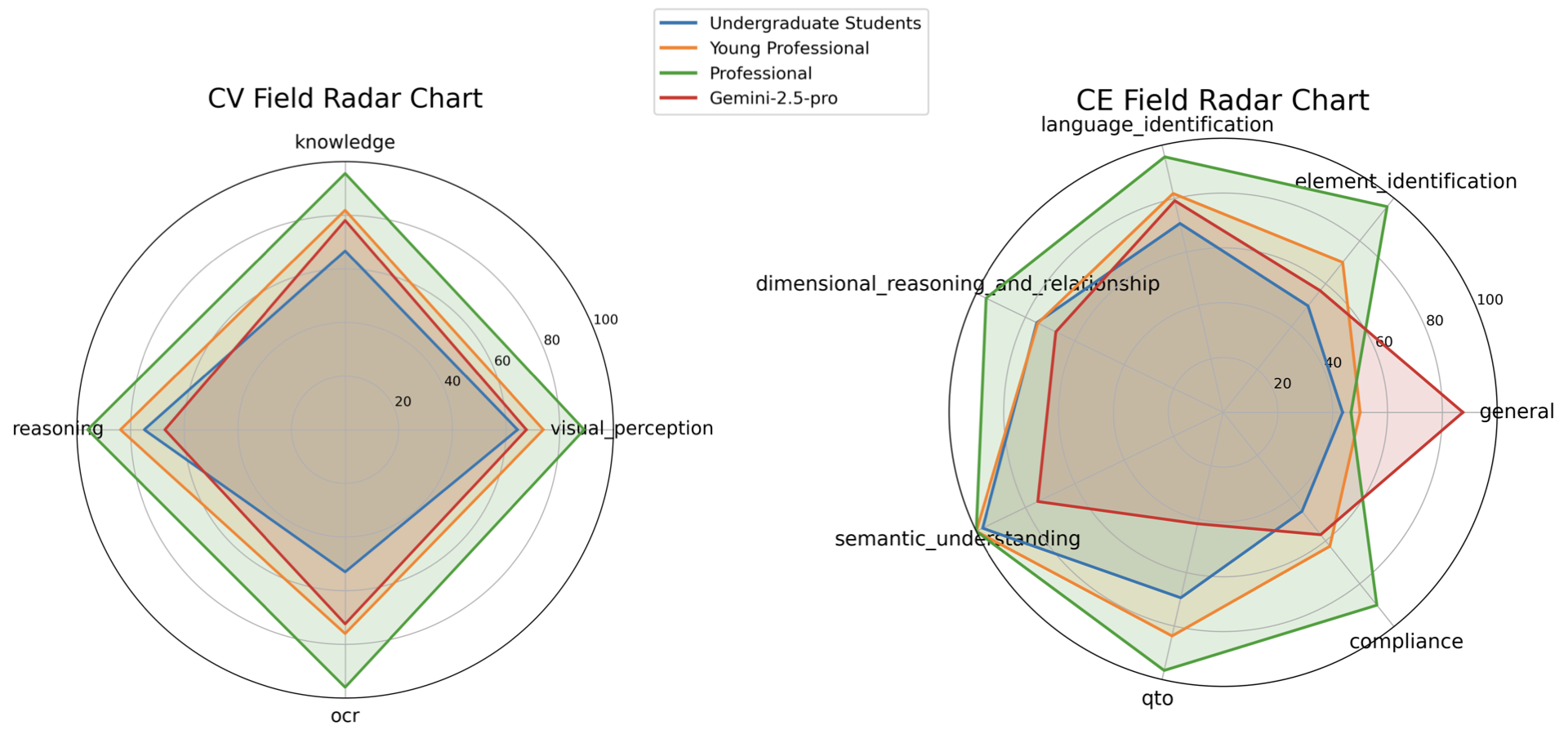}
    \caption{Comparison of multimodal reasoning performance across human cohorts and Gemini-2.5-pro on \OurBenchmark. Left: CV-derived competencies, Right: CE-derived competencies.}
    \label{fig:cecv}\vspace{-5pt}
\end{figure}

\paragraph{Reasoning Depth Analysis.}
Our Reasoning Depth category evaluates a model's ability to progress from basic perception to deep expertise. \textit{R1 (Visual Perception)} tests foundational skills, such as recognizing elements, where models like \texttt{Gemini-2.5-pro} (80.6) and \texttt{Gemini-2.5-flash} (77.4) score highest. This is followed by \textbf{R2 (Contextual Understanding)}, which requires connecting multiple pieces of information. Performance here shows a mixed pattern: while top-performing models (\texttt{Gemini-2.5-pro}, dropping to 76.5) and all human cohorts show a slight decrease, indicating the inherent difficulty of synthesis, several other models (e.g., \texttt{Qwen3-VL-8B} from 51.6 to 61.8) actually improve, suggesting they gain better insights as more context is provided. Finally, \textit{R3 (Expert Reasoning)}, which demands deep domain-specific inference, is clearly the most significant bottleneck. Scores for all models fall dramatically. This contrasts sharply with human \textit{Professionals}, whose R3 score (93.3) is their highest, highlighting the primary gap between current AI and true domain expertise.

\noindent\textbf{General MLLM aspects. }
Analyzing the ``MLLM Dimensions" reveals a distinct \textit{specialist} profile for models compared to humans. Models consistently excel in the \textbf{Knowledge (K)} dimension, with \texttt{Gemini-2.5-flash} (81.2) and \texttt{Gemini-2.5-pro} (78.1) achieving the highest scores. This suggests models are highly effective at retrieving and applying their vast store of accumulated knowledge. Conversely, humans demonstrate a clear advantage in the \textbf{Reasoning (R)} dimension, with the average human scoring 84.9 and professionals reaching 96.0. As illustrated in Figure~\ref{fig:cecv}, even undergraduate students are better than the state-of-the-art MLLM on reasoning capability. It highlights that humans retain a superior ability to perform flexible and abstract reasoning on complex visual information.

\noindent\textbf{Fine-Grained Construction Engineering Domain Aspects.}
Unlike the well-rounded human experts, Figure~\ref{fig:cecv} reveals that MMLM models exhibit a different performance pattern (i.e., ``spiky" specialist profile). MMLM strengths lie in \textit{Administration (Adm.)} and \textit{Language Understanding (LU)}, with \texttt{Gemini-2.5-pro} achieving 87.5. This indicates a strong grasp of domain-specific terminology and the ability to process administrative information from documents. However, the most significant weakness for all MLLMs is \textit{Quantity Take Off (QTO)}. With the top model (\texttt{Gemini-2.5-pro}) scoring only 41.7 and several models near 0.0, this task remains a major challenge. This difficulty is likely attributable to MLLMs' limitations in precise spatial grounding and the inability to accurately decompose and count objects in dense, large-scale engineering drawings. Domain-specific tasks such as quantity takeoff require multiple MLLM competencies simultaneously, including visual reasoning, OCR, and embedded domain knowledge, as illustrated in Figure~\ref{fig:dual}. Among open-source models, \texttt{Qwen3-VL-8B-Instruct} shows a very strong and competitive performance at 53.3, notably outperforming several proprietary models and showing exceptional strength in \textit{General Administration} aspect (81.2). More details are analyzed in Supplementary.

\textbf{Human Evaluation} The ``Construction Eng. Domain Aspects" category provides the most detailed insights. Human performance clearly illustrates the value of experience: scores show a progressive trend from \texttt{Undergraduate} (62.8) to \texttt{Graduate \& Young Professionals} (78.2) to \texttt{Professionals} (94.9). This trend is especially pronounced in the \textit{Compliance (Comp)} task, where the professional score (90.0) is nearly double that of undergraduates (46.2). This suggests that understanding complex specifications and requirements is a skill acquired through practical experience beyond academic achievement. In contrast, \textit{Semantic Understanding (SU)} appears to be a more fundamental skill, with all human cohorts performing exceptionally well (97.4-100.0).


\textbf{Content Realism.}
We also asked participants to rate question realism on a 1-5 scale (5=very realistic). The feedback confirmed our dataset's authenticity: a combined \textbf{71.2\%} of participants rated the questions as highly realistic (32.7\% rated '4'; 38.5\% rated '5'). Critically, 0\% of participants gave a '1' or '2' rating. This feedback validates \OurBenchmark as an authentic benchmark for real-world AEC challenges.

\textbf{Validating Multimodal Dependency (Text-Only Ablation)}
To validate that \OurBenchmark requires genuine multimodal reasoning and is not solvable through linguistic priors or context leakage, we performed a text-only ablation study. As shown in Table~\ref{tab:text_only_ablation}, this ablation is expected to cause a significant performance collapse, dropping scores to near or even lower than the random-guess baseline. This confirms that the visual information in the drawings is not just supplementary but \textit{indispensable} for solving the tasks, validating our dataset's reliance on grounded visual understanding.

\begin{table}[ht]
\centering
\caption{Text-Only Ablation Study}
\label{tab:text_only_ablation}
\begin{adjustbox}{width=0.45\textwidth,totalheight=0.9\textheight,keepaspectratio}
\begin{tabular}{lccc}
\toprule
\textbf{Model} & \textbf{Multimodal (Image+Text)} & \textbf{Text-Only (Ablation)} & \textbf{Drop ($\Delta$)} \\
\midrule
o3  & 58.7\% & 29.5\% & 29.2\% ($\downarrow$) \\
Gemini-2.5-flash & 66.3\% & 9.8\% & 56.5\% ($\downarrow$) \\
\midrule
Random (Baseline) & 27.2\% & - & - \\
\bottomrule
\end{tabular}
\end{adjustbox}\vspace{-5pt}
\end{table}



\section{Error Analysis}
\textbf{Overall.} As part of the analysis to understand why models made certain decisions for questions that they got wrong, their explanation or ``chain of thoughts" were investigated to break apart the steps it took, and where it may have gone wrong. A more detailed breakdown of these analysis will be provided in the Supplementary material.

\textbf{Visual Grounding and Perception Failures.}
A common failure mode occurs when the model's reasoning plan is logically sound, but its visual perception fails. This was most pronounced in \textbf{QTO (Quantity Take Off)} tasks. For instance, when asked to count columns along a specific grid line (e.g., "grid line Q"), the model's CoT would correctly state the plan: \textit{``Step 1: Locate Vertical Grid Line Q... Step 3: Trace Grid Line Q... Step 4: Identify and Count the Columns"}. However, the final count would be incorrect because the model visually ``derailed" while tracing, misidentifying the line or ``hallucinating" intersections that were not present.

This decoupling of logic and perception was also seen in detail-finding tasks. When asked to find a callout for a footing \textit{next to} another, the model's CoT would flawlessly describe finding a callout symbol (e.g., ``9 / S3.00"), but it would be the \textit{wrong} callout, having visually latched onto a different, nearby footing. This shows a fundamental inability to maintain precise spatial grounding in dense, symbol-rich IFC drawings illustrated in Figure~\ref{fig:overview}.


\textbf{Failure to Interpret Core Symbolic Language for Cross-Referencing.}
The most consequential breakdowns occur in tasks that require multi-step visual reasoning—particularly those involving callouts, cross-sheet links, or geometric dependencies. These tasks form the backbone of dimensional and compliance verification. When a model must search across multiple details for a governing reference, it often stops after identifying the first plausible link and fails to perform a complete visual scan. Once anchored to that partial discovery, the model shows a tendency to force-fit its reasoning to the available answer choices rather than reassessing the drawing. For engineering workflows—where reliability depends on accurately tracing references, elevations, sections, governing geometry, or scope boundaries—this behavior makes current models fundamentally untrustworthy for tasks requiring disciplined cross-referencing or multi-step interpretation.


\textbf{Lack of Expert Knowledge.}
A recurring source of failure is the MLLM’s limited domain knowledge in core construction and engineering disciplines. Models frequently misinterpret basic drafting conventions—such as weld symbols, standard structural symbols, and annotations with domain acronyms—because they lack the expert familiarity required to decode these symbols reliably. When MLLM cannot recognize what a triangular weld symbol means, or cannot distinguish a footing from other structural elements, it begins reasoning from a fundamentally incorrect premise. This knowledge gap does not merely produce isolated mistakes; it accelerates cascading errors, where misidentified elements lead to incorrect dimensional logic, false assumptions about load paths, or fabricated interpretations of the drawing. Without this baseline professional literacy, the model cannot properly read or interpret construction documents, and its reasoning degrades before the analytical process even begins.
\section{Conclusion}
We introduce \OurBenchmark, the first VQA benchmark built on complex, real-world IFC drawings, designed to assess the multimodal reasoning capabilities of MLLMs on tasks that reflect authentic engineering practice. We employ a dual categorization analysis that spans both MLLM capabilities and construction-engineering aspects. We hope this benchmark provides a foundation for evaluating AI systems in expert-driven domains. Future direction includes exploration of domain-specific steering, the role of model scale, the text and vision embedding capacity, and expansion into Civil, and Mechanical, Electrical, and Plumbing disciplines.


{
    \small
    \bibliographystyle{ieeenat_fullname}
    \bibliography{main}
}


\clearpage
\onecolumn

\setcounter{page}{1}          
\renewcommand{\thepage}{S\arabic{page}}  

       
\begin{center}
    {\Large\bfseries \OurBenchmark: A Real-World Benchmark for Multi-Depth Visual--Textual Reasoning on Construction Drawings}\\[1em]
    {\large\bfseries Supplementary Material}
\end{center}
\vspace{1em}

\tableofcontents



\appendix

\newpage
\section{Full Main Results}

Table~\ref{tab:full_results} presents the comprehensive evaluation of all models across three reasoning depths (R1--R3), four MLLM dimensions, and seven construction-specific domains. 

\begin{itemize}
    \item \textbf{Gaps Between Theory and Practice}: Previous studies have demonstrated that models like GPT-4o pass construction certification exams with nearly 90\% accuracy~\cite{CMLLM}. However, our results show a significant performance drop when these models are tasked with deciphering engineering drawings. For instance, GPT-4o achieves 48.9\% on \OurBenchmark. This highlights that while current MLLMs possess strong domain knowledge retrieval capabilities, they fundamentally lack the spatial and semantic visual reasoning required to execute real-world engineering workflows. This discrepancy aligns common notion that MLLMs excel at \textit{Knowledge} (their strongest dimension), rather than the visual reasoning required to execute real-world engineering workflows.
    \item \textbf{Multimodal Architecture}: Among the evaluated architectures, Gemini and Qwen family achieve state-of-the-art performance. This suggests that the native multimodal architecture of the family may have better visual grounding and model architecture for dense technical documents compared to others. Especially, \texttt{Gemini-3-pro-preview}\footnote{This was released on 11/19/2025 at the time of writing supplementary materials.}
 shows improved knowledge and OCR capability in their modality, leading to the best result (77.2\%), which now moves much closer to the graduate and young professionals' benchmark score. 
    \item \textbf{Benchmarking Against Human Expertise}: Most MLLMs significantly lag behind Graduate \& Young professionals (78.2\%) and the Professional Expert baseline (94.9\%). This gap between the best AI and industry experts indicates that while MLLMs are approaching the competency of an entry-level engineer, they are not yet reliable enough for autonomous professional practice.
\end{itemize}

Detailed model size impacts and qualitative error analysis are provided in the subsequent sections.

\begin{table*}[h]
\centering
\caption{Full Main Results. \textbf{MLLM Dimensions} (VP: Visual Perception, K: Knowledge, R: Reasoning, OCR: Optical Character Recognition); 
\textbf{Construction Eng. Domain Aspects} (Adm.: Administration, EI: Element Identification, LU: Domain Language Understanding, DR: Dimensional Recognition, SU: Design Semantic Understanding, QTO: Quantity Take Off, Comp.: Compliance). Underscored values indicate the highest score for each model within the MLLM Dimensions and CE Domain Aspects categories.}
\label{tab:full_results}
\begin{adjustbox}{width=1\textwidth,totalheight=0.9\textheight,keepaspectratio}
\begin{tabular}{lc|ccc|cccc|ccccccc}
\toprule
&& \multicolumn{3}{c|}{\textbf{Reasoning Depth}} & \multicolumn{4}{c|}{\textbf{MLLM Dimensions}} & \multicolumn{7}{c}{\textbf{Construction Eng. Domain Aspects}} \\
\midrule
\textbf{Model} & \textbf{Overall} & \textbf{R1} & \textbf{R2} & \textbf{R3} & \textbf{VP} & \textbf{K} & \textbf{R} & \textbf{OCR} & \textbf{Adm.} & \textbf{EI} & \textbf{LU} & \textbf{DR} & \textbf{SU} & \textbf{QTO} & \textbf{Comp} \\
\midrule


GPT-4o & 48.9 & 61.3 & 44.1 & 40.7 & 47.1 & \underline{59.4} & 44.3 & 48.7 & 56.3 & 32.4 & \underline{70.8} & 41.9 & 50.0 & 25.0 & 42.9 \\
o3 & 58.7 & 64.5 & 61.8 & 48.2 & 58.8 & \underline{65.6} & 54.1 & 59.2 & 62.5 & 56.8 & 54.2 & 48.4 & 61.5 & 25.0 & 42.9\\
\textbf{Gemini-2.5-pro} & \textbf{71.7} & 80.6 & 76.5 & 55.6 & 67.6 & \underline{78.1} & 67.2 & 72.4 & \underline{87.5} & 56.8 & 79.2 & 67.7 & 75.0 & 41.7 & 57.1\\
Gemini-2.5-flash & 66.3 & 77.4 & 73.5 & 44.4 & 64.7 & \underline{81.2} & 57.4 & 63.2 & \underline{87.5} & 51.4 & 79.2 & 51.6 & 67.3 & 33.3 & 71.4\\
Claude-4.5-Sonnet & 57.6 & 64.5 & 64.7 & 40.7 & 58.8 & \underline{59.4} & 57.9 & 54.1 & \underline{68.8} & 48.7 & 66.7 & 41.9 & 57.7 & 41.7 & 57.1\\
Claude-4.5-Haiku & 54.3 & 51.6  & 73.5  & 33.3  & 52.9  & \underline{65.6}  &  52.5 & 54.0  & \underline{62.5}  & 43.2 &  50.0 & 48.4 & 51.9  & 33.3   &42.9 \\
\midrule

LLaVA-OneVision-1.5-8B-Instruct \cite{OneVision-1.5} & 40.2 & 61.3 & 41.2 & 14.8 & 38.2 & \underline{46.9} & 31.1 & 39.5 & 43.8 & 24.3 & \underline{62.5} & 25.8 & 42.3 & 0.0 & 0.0\\
LLaVA-v1.6-Mistral-7B \cite{liu2023improved} & 28.3 & 25.8 & 29.4 & 29.6 & 29.4 & \underline{37.5} & 27.9 & 23.7 & 31.2 & 24.3 & 16.7 & \underline{32.3} & 30.8 & 25.0 & 14.3\\
LLaVA-v1.6-34B \cite{liu2023improved} & 42.4& 38.7&44.1 & 44.4 & 50.0 & \underline{53.1} & 42.6 & 40.8& \underline{43.8} & 37.8 & 41.7 & 35.5 & 40.4 & 33.3 & 42.9\\
Llama-3.2-11B-Vision-Instruct \cite{Llama32} & 39.1 & 45.2 & 47.1 & 22.2 & \underline{41.2} & 40.6 & 34.4 & 38.2 & \underline{50.0} & 29.7 & 45.8 & 32.3 & 42.3 & 8.3 & 28.6\\
Llama-4-Scout-17B-16E-Instruct \cite{meta2025llama4} & 53.3 & 58.1 & 58.8& 40.7 & 52.9 & \underline{62.5} & 52.5 & 55.3 & \underline{62.5} & 37.8 & 58.3 & 41.9 & 48.1 & 50.0 & 42.9 \\
Qwen3-VL-8B-Instruct \cite{qwen3technicalreport} & 53.3 & 51.6 & 61.8 & 44.4 & 50.0 & 53.1 & 45.9 & \underline{55.3} & \underline{81.2}& 40.5 & 66.7 & 45.2 & 46.2 & 33.3 & 42.9\\
Qwen3-VL-32B-Instruct \cite{qwen3technicalreport} & 58.7 & 48.4 & 79.4 & 44.4 & 55.9 & 56.3 & 55.7 & \underline{63.2} & \underline{75.0}&48.7 & 66.7 & 67.7 & 57.7 & 33.3 & 42.9  \\
Qwen3-VL-30B-A3B-Instruct \cite{qwen3technicalreport} & 47.8 & 51.6 & 58.8 & 29.6 & 38.2 & \underline{53.1}& 41.0 & 48.7 & \underline{62.5} & 29.7 & \underline{62.5}& 45.2 & 48.1 & 16.7 & 28.6 \\
InternVL3\_5-8B\cite{wang2025internvl3_5} & 50.0 & 58.1 & 58.8 & 29.6 & 44.1 & \underline{53.1} & 49.2 & 50.0 & 56.3 & 37.8 &\underline{58.3} & 41.9 & 38.5 & \underline{58.3} & 28.6\\
InternVL3.5-30B-A3B\cite{wang2025internvl3_5} & 41.3 & 51.6 & 35.3 & 37.0 & 26.5 & \underline{43.8} &  42.6 & 42.1 & 50.0 &  40.5 & \underline{45.8} & 45.1 & 46.1 & 8.3 & 14.3 \\
InternVL3\_5-38B\cite{wang2025internvl3_5} & 50.0 & 48.4 & 64.7 & 33.3 & 41.2 & \underline{59.4} & 47.5 & 54.0 & \underline{62.5} & 35.1 & \underline{62.5} & 54.8 & 44.2 & 16.7 &14.3  \\
Phi-4-multimodal-instruct  \cite{microsoft2025phi4} & 40.2 & 41.9 & 47.1 & 29.6 & 29.4 & 40.6 & \underline{42.6} & 40.8 & 37.5 & 40.5 & \underline{45.8} & 35.5 & 40.4 & 25.0 & 28.6\\

\midrule

Random & 27.2 & 25.8 & 17.7 &  40.7 & 23.5 & \underline{37.5} &  27.9 & 29.0 & 37.5 &  18.9 & 29.2 & 29.0 &  23.0 & 16.6 & \underline{42.8} \\
Human & 68.4 & 75.6 & 72.6 & 59.9 & 75.8 & 81.4 & \underline{84.9} & 75.0 & 46.8 & 71.9 & 82.7 & 82.1 & \underline{99.2} & 83.3 & 66.2\\
\hspace{1em} * Undergraduate & 62.8 & 71.8 & 70.2 & 53.8 & 64.3 & 66.7 & \underline{74.9} & 53.0 & 43.6 & 49.7 & 70.7 & 75.4 & \underline{97.4} & 69.4 & 46.2\\
\hspace{1em} * Graduate \& Young Professionals & 78.2 & 85.4 & 76.6 & 68.8 & 73.9 & 81.9 & \underline{83.8} & 76.0 & 50.0 & 70.0 & 81.9 & 75.0 & \underline{100.0} & 83.7 & 62.5\\
\hspace{1em} * Professionals & \textbf{94.9} & 90.0 & 85.0 & 93.3 & 89.1 & 95.6 & \underline{96.0} & \underline{96.0} & 46.7 & 96.0 & 95.6 & 96.0 &\underline{100.0} & 96.6 & 90.0\\
\midrule
\textbf{Gemini-3-pro-preview} & \textbf{77.2} & 80.6 & 79.4 & 70.4 & 76.5 & \underline{81.3} & 72.1 & 80.3 & \underline{93.8} & 75.7 & 83.3 & 77.4 & 78.8 & 33.3 & 85.7\\
\bottomrule
\end{tabular}
\end{adjustbox}
\end{table*}

\section{Author contributions}

\begin{itemize}
    \item \textbf{Project Conception:} Y.J., J.F., M.G.
    \item \textbf{Task Definition \& Benchmark Design:} Y.J., J.F., M.G.
    \item \textbf{Data Curation \& Management:} Y.J., J.F., M.G.
    \item \textbf{Benchmark Question Generation:} Y.J., J.F., M.G.
    \item \textbf{Model Evaluation:} Y.J., J.F.
    \item \textbf{Qualitative \& Quantitative Analysis:} Y.J., J.F.
    \item \textbf{Writing \& Visualization:} Y.J., J.F.
    \item \textbf{Supervision:} Y.J., J.F., M.G.
\end{itemize}

\section{Limitations and future work}

\begin{itemize}

\item \textbf{Dataset Scale, Density, and the Expert Bottleneck}
Our final dataset consists of 92 high-quality, professionally annotated samples. We acknowledge that this scale is smaller than general-domain VQA benchmarks or DesignQA~\cite{designqa}. However, as highlighted in comparisons (in the main paper), there is a distinct trade-off between dataset size and reasoning depth. High-level cognitive tasks especially toward real-world engineering workflows require annotations from licensed experts, creating an ``expert bottleneck'' that limits rapid scaling.

Crucially, despite the modest total count, \OurBenchmark is larger and more diverse than the construction-engineering subsets of existing generalist benchmarks. As shown in Figure 4 in the main paper, the relevant subsets of MM-Vet (4 samples), ScienceQA (24 samples), and RBench-M (54 samples) are significantly smaller. Furthermore, where other datasets typically focus on a single question type (e.g., from textbook, exam), \OurBenchmark spans 7 distinct construction-engineering workflow aspects. While smaller in volume, our dataset offers a higher ``information density'' and a more comprehensive evaluation of professional versatility than existing alternatives.

\item \textbf{Closed vs. Open-Ended}
To ensure robust and reproducible comparisons, the majority of our quantitative evaluation relies on Multiple-Choice Questions. We recognize, however, that real-world engineering workflows operate in an open-ended setting—practitioners do not select from a list of options but must generate solutions derived from compliance checks to spatial and quantity calculations. 
While we included an open-ended subset in our benchmark, it remains a challenge. Future iterations of \OurBenchmark aim to establish a rigorous question set for open-ended engineering VQA.

\item \textbf{Scope of Engineering Disciplines}
\OurBenchmark currently emphasizes structural engineering and construction management, utilizing real-world IFC-level structural drawings. While structural discipline is the backbone of any construction, the domain is multidisciplinary. Our current scope excludes other disciplines such as Architectural, Civil, and other critical system disciplines such as Mechanical, Electrical, and Plumbing (MEP).  While the way to read IFC drawings and the main rationale are similar across disciplines, visual perception and the reasoning process differ. MEP drawings rely heavily on schematic symbols, deciphering complex relations and connections, rather than the physical scale and dimensioning critical to structural drawings.  Future work must expand the dataset to cover these modalities. This would allow for a more holistic assessment of an MLLM's ability to understand construction engineering drawings in a meaningful way.
\end{itemize}

\section{Benchmark details}

\subsection{Accessing \OurBenchmark}
\OurBenchmark is an expertly curated benchmark datasets that utilizes Issued for Construction (IFC) level drawings. In this iteration of \OurBenchmark, the focus was made on structural discipline drawings, as structural drawings include the backbone of any building structures that become the basis for coordination with other disciplines. 

The QA sets within the drawings focus on practical questions that would be asked by construction industry professionals to better understand design intent and design requirements. Contrary to many dataset that may require `intrinsic' or `latent' knowledge, \OurBenchmark focuses on reasoning within the given context window of a drawing or a portion of the drawing, to truly test MLLMs capability in understanding local contexts without hallucinating. 

The \OurBenchmark is shared on \url{https://huggingface.co/datasets/S2-MIND/DrawingVQA}. It is
distributed under the Creative Commons Attribution-NonCommercial-ShareAlike 4.0 International (CC BY-NC-SA 4.0) license. Please note that most of the drawings do contain private information such as the architects, engineers, and general contractors. Any information that indicates private information were redacted. 

\subsection{General Guidelines}
There are two types of questions asked within this QA set: Multiple-Choice Questions, and Open-Ended Questions. The multiple choice questions included True/False, and choices of more than 4 options with one correct option. The closed ended questions encompassed short descriptive responses, however, the questions were asked in a way to ensure that long answers were avoided. Each questions had at least one input of a construction drawing image, where it was either a full drawing image, or a proportion of the full drawing image. 

Construction drawings utilized in the industry are typically saved as PDFs for portability. For the purpose of benchmarking, these drawings were converted into PNG images. As part of a process to export from PDF to images, there are various resolutions that can be selected in the unit of pixels per inch (ppi). For the \OurBenchmark, range of resolution from 25 to 125 were extracted for testing. Practically, the resolutions can defer based on the usage, where for lightweight purpose, the ppi can be as low as 75 ppi, but in other cases such as for printing applications, this can be up until 300 ppi. 

The table below shows the average width and height of the images used, as it contains full drawing images to portion of the drawing image, to a very small patch for symbol detection. 

\subsection{Question curation guidelines}
To curate the dataset, 33 Issued for Construction level structural discipline drawings were collected. These drawings became the basis for brainstorming potential questions that can be asked, specifically in the sense that if this was a real project, what sort of questions would professionals ask first when seeing the drawings. These questions and answer pairs were recorded first to ensure that the complexity of the questions were realistic, and matched what was needed for a meaningful benchmark. 

Upon completion for the first round of Question-Answer creation, a more scrutinized evaluation was made to come up with plausible options that may arise as a result of misreading or misunderstanding the drawings. These options were added as part of the Question-Answer pair. Finally, these questions were evaluated and designated for its dual categorization to possibly create meaningful insights upon completion of the benchmark testing. 

\begin{table}[ht]
\centering
\caption{Drawing image size variations in pixels}
\label{tab:api_short}
\begin{adjustbox}{width=0.4\textwidth,totalheight=0.15\textheight,keepaspectratio}
\begin{tabular}{llll}
\toprule
\textbf{} & \textbf{Minimum} & \textbf{Median} & \textbf{Maximum}\\
\midrule
Width & \texttt{54} & \texttt{4064} & \texttt{4800}\\
Height & \texttt{82} & \texttt{3000}& \texttt{3600}\\
\bottomrule
\end{tabular}
\end{adjustbox}
\end{table}




\subsection{Dataset VQA Structure}

The code block below indicates the dataset schema and structure used to conduct the test in various models this work has benchmarked. 

\newenvironment{centeredlisting}
  {\par\begin{center}\begin{minipage}{0.9\linewidth}}
  {\end{minipage}\end{center}}
\begin{centeredlisting}
\begin{lstlisting}[style=clean, language=Python]
    {
        "id": Integer,
        "image_name": String,
        "image2_name": String,
        "question": String,
        "options": {
            "A": String,
            "B": String,
            "C": String,
            "D": String
        },
        "answer": String,
        "explanation": String,
        "cv_field": List[String]
        "cv_subfield": List[String]
        "ce_field": List[String]
        "ce_subfield": List[String],
        "topic_difficulty": String,
        "question_type": String
    },
\end{lstlisting}
\end{centeredlisting}


\subsection{Training contamination mitigation}

A critical challenge in benchmarking MLLMs is the risk of data contamination, where test samples inadvertently appear in the model's pre-training corpus. This leakage allows models to solve problems through memorization rather than reasoning. 

To guarantee the integrity of our evaluation, \OurBenchmark is designed to be strictly contamination-free through two key mechanisms:

\begin{itemize}
    \item \textbf{Proprietary and Offline Image Sources:} Unlike benchmarks derived from public internet crawls, the IFC construction engineering drawings in our dataset were sourced from private construction projects and have not been visible to public search engines.
    
    \item \textbf{Novel Annotation and Temporal Separation:} The Question-Answer (QA) pairs were generated \textit{de novo} by our research team in late 2025, following industry practices. This creation date post-dates the knowledge cutoff of all models tested. 
\end{itemize}

Consequently, we can confirm that \OurBenchmark represents a true zero-shot evaluation environment. The performance metrics reported in this study reflect the models' genuine capability to contextualize novel visual engineering context window of a drawing, rather than their ability to recall training data.

\subsection{Question Format: Multiple-Choice vs. Open-Ended}

To ensure rigorous and reproducible evaluation, \OurBenchmark employs a Multiple-Choice Question (MCQ) format. The distractors within these MCQs are meticulously expert-designed to reflect realistic cognitive and procedural errors commonly made by practitioners when interpreting IFC (Issued for Construction) drawings. This design ensures the benchmark serves as a reliable proxy for real-world engineering challenges.

To verify that the closed-ended format does not artificially alter the inherent difficulty of the tasks, we conducted an ablation study by converting a subset of 20 MCQs into open-ended questions. An evaluation of MLLM performance across these two formats yielded no statistically significant difference in accuracy (paired t-test, $p = 0.09$). This confirms that the MCQ format in \OurBenchmark effectively evaluates reasoning capabilities without introducing a format-based advantage or bias.

\subsection{Generalization to Unseen Drawing Styles}

We prioritize reasoning density over repetitive volume. To further validate the generalization capabilities of our dataset, we conducted an additional evaluation using unseen data. Specifically, we compiled a supplementary set of 49 new VQA pairs derived from 7 newly introduced drawings, distributing 7 QA sets evenly across 7 distinct Construction Engineering (CE) dimensions.

Comparing model performance on this unseen set against the original benchmark, a paired t-test revealed no statistically significant difference in overall MLLM accuracy ($p = 0.07$). This confirms that the performance benchmarked by our dataset remains consistent and generalizes effectively when models are confronted with different drawing styles, due to our drawing QA is careful engineering reasoning required not just extracting information.

\subsection{Ethical Consideration}

\begin{itemize}
    \item \textbf{Copyright and Licensing:} We maintain strict adherence to all applicable copyright and licensing regulations. 
    
    \item \textbf{Data Privacy and Anonymity:} All project-specific identifiers (e.g., client names, specific site addresses) were redacted prior to inclusion.
\end{itemize}


\section{Dual Categorization Taxonomy}

To move beyond aggregate accuracy metrics and diagnose specific model bottlenecks, we developed a dual-category taxonomy. This framework cross-references the \textit{Domain-Specific Dimension} (the engineering intent of the query) with the \textit{Cognitive Capability} (the underlying mechanism required by the MLLM to solve it).

The taxonomy classifies queries into seven domain dimensions, ranging from \textit{General Administrative} tasks to complex \textit{Code and Specification Compliance verification}. As detailed in Table~\ref{tab:dual_taxonomy}, each dimension is associated with the corresponding primary MLLM capabilities necessary for resolution:
\begin{itemize}
    \item \textbf{OCR (Text Regocnition/Understanding):} Essential for interpreting dense technical annotations, tables, and callouts (i.e., other cross-references).
    \item \textbf{Visual Perception:} Low-level identification of graphical entities (symbols, lines, shapes), delineation of major spaces within drawings such as sub-drawings or details, and object counting.
    \item \textbf{Reasoning:} Higher-order processing, including \texttt{Spatial} (2D-to-2/3D reasoning), \texttt{Alignment} (cross-referencing between visual and textual content), and \texttt{Visual} (compositional analysis) reasoning. An example can include understanding spatial composition of elements within the model (e.g. this element A is on the left of element B, element C exists between grid line D and E).
    \item \textbf{Knowledge:} Retrieval of external domain facts or latent knowledge (e.g., standard acronyms, drawing standards, industry best practice, specifications or other engineering standards) not explicitly visible in the pixel data.
\end{itemize}


\begin{table}[h!]
\centering
\caption{The \OurBenchmark Domain-Capability Map. We define seven domain dimensions and identify the primary MLLM cognitive capabilities typically required to solve tasks within each category.}
\label{tab:dual_taxonomy}
\begin{adjustbox}{width=\textwidth,keepaspectratio}
\begin{tabular}{l p{6cm} p{5cm}}
\toprule
\textbf{Domain-Specific Dimension} & \textbf{Task Description} & \textbf{Primary Cognitive Dependencies} \\
\midrule
\textbf{1. General Admin.} & Identifying project participants, sheet metadata, dates, and drawing types. & OCR, Visual Perception \\
\midrule
\textbf{2. Domain Element Identification} & Interpreting specific architectural, structural element symbols, callout symbols and other administrative components. & Visual Perception, Reasoning (Alignment, Visual), OCR \\
\midrule
\textbf{3. Discipline Language Understanding} & Identifying discipline-specific acronyms, abbreviations, and technical nomenclature. & OCR, Knowledge, Reasoning (Visual, Alignment)\\
\midrule
\textbf{4. Dimensional Understanding} & Mapping sections to plan views, locating intersections, and understanding spatial offsets. & Reasoning (Alignment, Spatial), OCR \\
\midrule
\textbf{5. Design Semantic Understanding} & Inferring coordination needs, distinguishing design alternatives, and interpreting context. & Knowledge, OCR, Reasoning (Visual, Spatial) \\
\midrule
\textbf{6. Quantity Take-Off (QTO)} & Counting elements, measuring linear/area quantities, and extracting categories from schedules. & Reasoning (Alignment), Visual Perception, OCR \\
\midrule
\textbf{7. Compliance \& Specs} & Matching components to specifications, identifying scope gaps, or verifying code compliance. & Knowledge,  Reasoning (Alignment, Visual, Spatial), OCR\\
\bottomrule
\end{tabular}
\end{adjustbox}
\end{table}

\section{Experiments details}

\subsection{Evaluation Prompts and Parsing}\label{sec:prompt}

To evaluate the model's performance, we structured the input prompts to encourage Chain-of-Thought (CoT) reasoning and specified a strict output format. Depending on the question type, we appended the following instructions to the input.
\vspace{2pt}

\noindent\textbf{Multiple Choice:}
\vspace{-3pt}\begin{tcolorbox}[colback=gray!5!white,colframe=black!70!white,arc=3mm]
The following is a multiple choice question. 

Think step by step and then output the answer in the format of $\backslash$``The answer is (X)$\backslash$" at the end.
\vspace{1em} 

\{\{[QUESTION]\}\}
\vspace{1em} 

Options:

\{\{[CHOICES]\}\}
\end{tcolorbox}
\vspace{2pt}

\noindent\textbf{Open-Ended:}
\vspace{-3pt}\begin{tcolorbox}[colback=gray!5!white,colframe=black!70!white,arc=3mm]
The following is an open-ended question (with an explicit numeric answer). 

Think step by step and then output the answer in the format of $\backslash$``The answer is (X)$\backslash$" at the end.
\vspace{1em} 

\{\{[QUESTION]\}\}
\vspace{1em}

\end{tcolorbox}



Despite these explicit formatting instructions, the models occasionally generated heterogeneous output formats. Common variations included:
\begin{itemize}
    \item Standard sentences: ``The answer is A.'' or ``The answer is (A)''
    \item LaTeX formatting: ``The final answer is \texttt{\textbackslash boxed\{A\}}.''
    \item Markdown emphasis: ``The answer is **A: XX**.''
    \item Minimalist output: ``A'' or ``A: XX''
\end{itemize}

To robustly extract the predicted label across these variations, we implemented a regular expression pattern designed to capture the answer key while ignoring surrounding formatting or punctuation. Other variations, except the above list, are not handled in the parsing process. 




\subsection{Model details}
Table~\ref{tab:api} details the specific versions and API endpoints utilized in our experiments to ensure full reproducibility of our results.

\begin{table}[ht]
\centering
\caption{MLLM API endpoints and sources.}
\label{tab:api}
\begin{adjustbox}{width=0.75\textwidth,totalheight=0.35\textheight,keepaspectratio}
\begin{tabular}{lll}
\toprule
\textbf{Model} & \textbf{API Endpoint} & \textbf{Source}\\
\midrule
GPT-4o & \texttt{gpt-4o-2024-08-06} & OpenAI API \\
o3  & \texttt{o3-2025-04-16} & OpenAI API \\
Gemini-2.5-pro & \texttt{gemini-2.5-flash} & Google Gemini API\\
Gemini-2.5-flash & \texttt{gemini-2.5-pro} & Google Gemini API\\
Claude-4.5-Sonnet & \texttt{claude-sonnet-4-5-20250929}& Claude API\\
Claude-4.5-Haiku & \texttt{claude-haiku-4-5-20251001} & Claude API\\
\midrule
LLaVA-OneVision-1.5-8B-Instruct & \texttt{lmms-lab/LLaVA-OneVision-1.5-8B-Instruct}& HuggingFace, local inference \\
LLaVA-v1.6-Mistral-7B & \texttt{llava-hf/llava-v1.6-mistral-7b-hf} & HuggingFace, local inference\\
LLaVA-v1.6-34B & \texttt{llava-hf/llava-v1.6-34b-hf} & HuggingFace, local inference\\
Llama-3.2-11B-Vision-Instruct & \texttt{meta-llama/Llama-3.2-11B-Vision-Instruct} & HuggingFace, local inference\\
Llama-4-Scout-17B-16E-Instruct & \texttt{meta-llama/Llama-4-Scout-17B-16E-Instruct} & HuggingFace, local inference\\
Qwen3-VL-8B-Instruct &\texttt{Qwen/Qwen3-VL-8B-Instruct} & HuggingFace, local inference\\
Qwen3-VL-32B-Instruct & \texttt{Qwen/Qwen3-VL-32B-Instruct} & HuggingFace, local inference\\
Qwen3-VL-30B-A3B-Instruct & \texttt{Qwen/Qwen3-VL-30B-A3B-Instruct}& HuggingFace, local inference\\
InternVL3.5-8B & \texttt{OpenGVLab/InternVL3\_5-8B-HF} & HuggingFace, local inference\\
InternVL3.5-38B & \texttt{OpenGVLab/InternVL3\_5-38B-HF} & HuggingFace, local inference\\
InternVL3.5-30B-A3B & \texttt{OpenGVLab/InternVL3\_5-30B-A3B-HF} & HuggingFace, local inference\\
Phi-4-multimodal-instruct  & \texttt{microsoft/Phi-4-multimodal-instruct} & HuggingFace, local inference \\

\bottomrule
\end{tabular}
\end{adjustbox}
\end{table}

\subsection{Human baseline}
As part of an effort to ensure that the tests being conducted on various SOTA MLLMs are reasonable and not superficial or impractical, a human baselines for the same questions were created to compare their performance against the latest models. 

Similar to how various models may have variant parameter sizes or access to different knowledge, a similar delineation was made to the human baseline tests based on their industry experience in years. We recruited 52 participants within the Civil Engineering or Construction Management industry with 3 levels of expertise in the following manner: Undergraduate (Years of experience = 0 years); Young Professionals such as graduate students or those who have just started working in the industry (0 $\leq$ Years of experience $\leq$ 2 years); Professionals (3 $\leq$ Years of Experience). This delineation will also aim to help create human benchmark checkpoints that are comparable with model performance. 

As many of the participants were students and working professionals, the questionnaire for the baselines was shortened to 20 questions per person, but the questions were asked in the exact same format as to the MLLMs, with one image + multiple choices for the closed-ended questions, and one image + free response for the open-ended questions. The following rules were used to set up the human benchmark test:

\begin{itemize}
    \item Questions were provided in a google form.
    \item Participants did not view any questions prior to completing the form. 
    \item Any internet, LLM or other resource access were prohibited, to preserve testing hygiene, where only the context window of the drawing image, as well as their prior experiences in the industry, are utilized. 
    \item Time limit was set up to be 20 minutes for 20 questions, to ensure all participants had the same control variable, regardless of their experiences.
\end{itemize}


At the end of the benchmark test, all the participants were asked to rank the difficulty of the QA set, as well as the realistic nature of the QA set to the practical industry application. The results collected by the participants are shown below. 

\begin{table}[ht]
\centering
\caption{DrawingVQA Human Benchmark Test Feedback Results (\%).}
\label{tab:api_human}
\begin{adjustbox}{width=0.6\textwidth,totalheight=0.15\textheight,keepaspectratio}
\begin{tabular}{llllll}
\toprule
\textbf{} & \textbf{1} & \textbf{2} & \textbf{3} & \textbf{4} & \textbf{5}\\
\midrule
How realistic were the questions? & \texttt{0.0} & \texttt{0.0} & \texttt{28.3} & \texttt{32.1} & \texttt{39.6} \\
How difficult were the questions? & \texttt{3.8} & \texttt{17.0} & \texttt{50.9} & \texttt{28.3} & \texttt{0.0} \\
\bottomrule
\end{tabular}
\end{adjustbox}
\end{table}

\begin{table}[ht]
\centering
\caption{Participant votes for the dataset that best requires reasoning skills on real-world AEC industry context and engineering practices.}
\label{tab:dataset_votes}
\begin{adjustbox}{width=0.65\textwidth,keepaspectratio}
\begin{tabular}{llcc}
\toprule
\textbf{Dataset} & \textbf{Source} & \textbf{Votes} & \textbf{Percentage (\%)} \\
\midrule
DrawingVQA & Construction drawings from real-world projects & 40 & 76.9 \\
MMMU       & FE exam or college exam                      & 7  & 13.5 \\
CEQuest    & Textbooks and guidelines                     & 3  & 5.8  \\
None       & -                                            & 2  & 3.8  \\
\bottomrule
\end{tabular}
\end{adjustbox}
\end{table}

\subsection{Dense and MoE models}

To understand the trade-off between inference efficiency and reasoning accuracy on IFC construction drawings, we compared Dense architectures against Mixture-of-Experts (MoE) variants within the same model families. MoE models are designed to reduce computational cost by activating only a subset of parameters per token.

Table~\ref{tab:moe_vs_dense} summarizes the performance across the Qwen, InternVL, Llama, and LLaVA families. We observe a distinct trend where \textit{Dense architectures generally outperform their MoE counterparts on the DrawingVQA benchmark}.

\begin{itemize}
    \item \textbf{Qwen Family:} The Qwen3-VL-8B (Dense) achieved \textbf{53.3\%} accuracy, surpassing the Qwen3-VL-30B-A3B (MoE) which scored \textbf{47.8\%}, despite the MoE model having nearly $4\times$ the total parameters.
    \item \textbf{InternVL Family:} A similar pattern emerges, where the Dense 8B model (50.0\%) significantly outperforms the MoE 30B variant (41.3\%).
    \item \textbf{Scaling Behavior:} While the Llama-4-Scout MoE achieved a competitive 53.3\%, it required a massive total parameter budget of 109B to match the performance of the significantly smaller Qwen-8B Dense model.
\end{itemize}

These results suggest that for high-fidelity visual tasks requiring global context interpretation—such as reading complex construction drawings—the sparsity of MoE models may be a limiting factor compared to the dense connectivity of standard Transformers.

\begin{table}[h]
\centering
\caption{Performance comparison between Dense and Mixture-of-Experts (MoE) models across different parameter scales. $\dag$ These numbers are from each model's report and official website.}
\label{tab:moe_vs_dense}
\resizebox{0.65\linewidth}{!}{%
\begin{tabular}{llccc}
\toprule
\textbf{Model} & \textbf{Arch.} & \textbf{Total Params.$\dag$} & \textbf{Activated Params.$\dag$} & \textbf{Acc. (\%)} \\
\midrule
\textit{Qwen Family}  \cite{qwen3technicalreport} & & & & \\
Qwen3-VL-8B-Instruct  & Dense & 8B & 8B & 53.3 \\
Qwen3-VL-32B-Instruct  & Dense & 32B & 32B & 58.7 \\
Qwen3-VL-30B-A3B-Instruct & MoE & 30B & 3B & 47.8 \\
\midrule
\textit{InternVL Family} \cite{wang2025internvl3_5}& & & & \\
InternVL3.5-8B  & Dense & 8B & 8B & 50.0 \\
InternVL3.5-38B  & Dense & 38B & 38B & 50.0 \\
InternVL3.5-30B-A3B  & MoE & 30B & 3B & 41.3 \\
\midrule
\textit{Llama Family} \cite{Llama32} & & & & \\
Llama-3.2-11B-Vision-Instruct  & Dense & 11B & 11B & 39.1 \\
Llama-4-Scout-17B-16E-Instruct & MoE & 109B & 17B & 53.3 \\
\midrule
\textit{LLaVa Family} \cite{liu2023improved} & & \\
LLaVA-v1.6-Mistral-7B & Dense & 7B & 7B& 28.3 \\
LLaVA-v1.6-34B& Dense & 34B& 34B& 42.4\\
\bottomrule
\end{tabular}%
}
\end{table}

\subsection{No-Image Ablations}

To quantify the extent to which models rely on visual information versus language priors, we conducted a no-image ablation study, a standard diagnostic in VQA evaluations~\cite{microvqa,mmmu}. In this setting, the MLLMs are provided with the textual question alone, without the accompanying visual context. Following \cite{microvqa}, we prepended the prompt with the following sentence:

\begin{quote}
\texttt{If an image is mentioned, ignore this information and try your best to answer the question.}
\end{quote}

A rigorous VQA benchmark is ``vision-centric'', ensuring that questions cannot be solved solely through common sense reasoning or textual artifacts. \OurBenchmark was explicitly curated to reflect realistic engineering problems encountered by professionals on job sites, which inherently require visual verification.

As presented in Table~\ref{tab:text_only_ablation_all}, the significant performance drop ($\Delta$) observed across high-performing models provides strong empirical evidence that \OurBenchmark is vision-dependent. Qualitative analysis reveals that when text-only models answer correctly, they often rely on hallucinations (inventing scenarios), bias toward specific answers (e.g., defaulting to ``True'' on binary questions), or exploit domain priors (e.g., guessing typical quantities of drawing notes). However, the substantial performance gap (around or even below that random guess) confirms that reasoning on this benchmark necessitates visual grounding.

Notably, refusal rates offer insight into model safety and grounding capabilities. Models such as \texttt{Gemini-2.5-flash} and \texttt{LLaVA-OneVision} exhibited high refusal rates (55.4\% and 81.5\%, respectively), correctly identifying that the questions were unanswerable without the visual drawing context.

\begin{table}[h!]
\centering
\caption{Text-Only Ablation Study. Value is a percentage.}
\label{tab:text_only_ablation_all}
\begin{adjustbox}{width=0.85\textwidth,totalheight=0.9\textheight,keepaspectratio}
\begin{tabular}{lcccc}
\toprule
\textbf{Model} & \textbf{Multimodal Acc. (Image+Text)} & \textbf{Text-Only Acc. (Ablation)} & \textbf{Drop ($\Delta$)} & \textbf{Refuse to answer (Require image)}  \\
\midrule
GPT-4o & 48.9 & 31.5 & 17.4 ($\downarrow$) & 15.2 \\
o3  & 58.7  & 29.5 & 29.2 ($\downarrow$)&   1.1\\
Gemini-2.5-pro & \textbf{71.7} & 37.0 & 34.7 ($\downarrow$) & 13.0 \\
Gemini-2.5-flash & 66.3 & 9.8 & 56.5 ($\downarrow$) & 55.4 \\
Claude-4.5-Sonnet & 57.6 & 35.9 & 21.7 ($\downarrow$) & 2.2\\
Claude-4.5-Haiku & 54.3  &  18.5 & 35.8 ($\downarrow$) & 56.5\\
\midrule
LLaVA-OneVision-1.5-8B-Instruct \cite{OneVision-1.5} & 40.2 & 8.7 & 31.5 ($\downarrow$) & 81.5\\
LLaVA-v1.6-Mistral-7B \cite{liu2023improved}& 28.3 & 25.0 & 3.3($\downarrow$) & 2.2 \\
Llama-3.2-11B-Vision-Instruct \cite{Llama32} & 39.1 & 30.4 & 8.7($\downarrow$) & 3.3  \\
Qwen3-VL-8B-Instruct \cite{qwen3technicalreport} & 53.3 & 34.8 & 18.5 ($\downarrow$) & 9.9 \\
InternVL3.5-30B-A3B\cite{wang2025internvl3_5} &41.3 & 37.0 & 4.3 ($\downarrow$) & 3.3 \\
Phi-4-multimodal-instruct \cite{microsoft2025phi4} & 40.2&31.5 &8.7 ($\downarrow$) & 17.4 \\

\midrule
Random (Baseline) & 27.2 & - & - & - \\
\bottomrule
\end{tabular}
\end{adjustbox}
\end{table}

\subsection{Breakdown results on Dual Category Mapping}

\begin{figure}[h]
    \centering
    \includegraphics[width=0.70\linewidth]{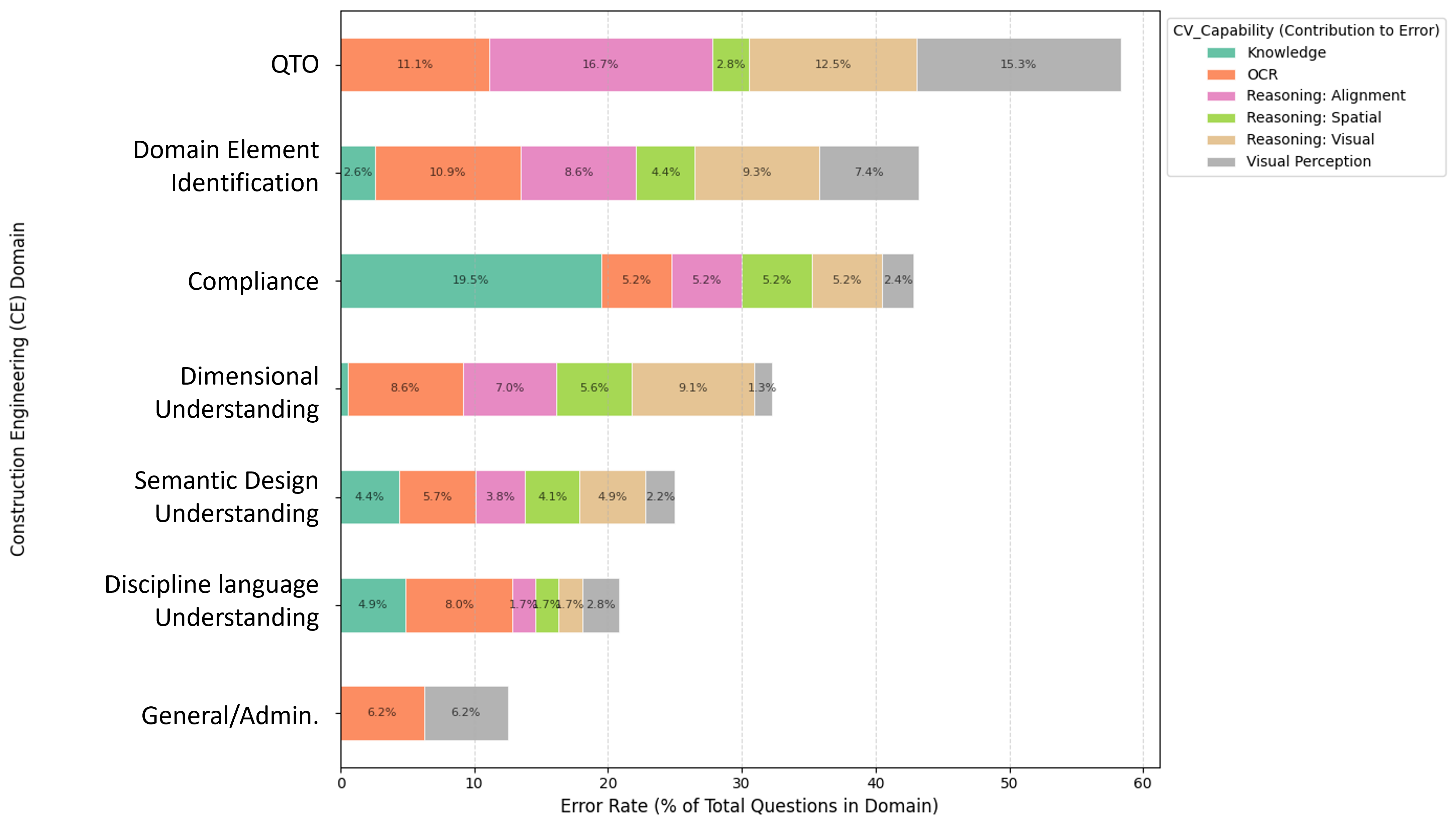}
    \caption{\textbf{Dual-Category Error Breakdown.}}
\label{fig:error_breakdown}
\end{figure}

To identify specific cognitive bottlenecks in construction drawing interpretation, we introduce our dual-category mapping to the error instances of the top-performing model, Gemini-2.5-Pro. Figure~\ref{fig:error_breakdown} illustrates the distribution of failure causes across seven domain-specific tasks. The analysis reveals that failure modes are highly context-dependent. General administrative tasks (``Admin.'') require \textit{OCR} and \textit{Visual Perception} capabilities from MLLMs, indicating the model gets answers by reading explicit information in drawings. In contrast, technical tasks such as ``Domain Element Identification'' and ``Dimensional Understanding (Cross-Referencing)'' exhibit a shift toward \textit{Reasoning} (spatial, visual, and alignment), OCR, and \textit{Visual Perception} capabilities required. ``QTO'' (Quantity Take-off) shows sensitivity to \textit{OCR} and \textit{Visual alignment} errors, highlighting the challenge of accurately grounding text within complex graphical geometries. On the other hand, ``Compliance" requires knowledge beyond explicit information within the drawings. This heterogeneity indicates that a single optimization strategy is insufficient; distinct construction tasks require targeted improvements in specific multimodal capabilities.

\subsection{More experiments on Image resolution}

We investigate the impact of image resolution on the performance of MLLMs within the context of engineering drawings. In our dataset construction, we distinguish between Region of Interest (ROI) images and Full Drawing Sheets. While ROI images are typically user-generated crops that do not require standardization, the Full Drawing Sheet serves as the global context and must be rasterized at a resolution that balances legibility with computational efficiency.

To determine the optimal setting, we generated variants of the full drawing sheets at pixel densities ranging from 25 PPI to 125 PPI. Table~\ref{tab:image_dimensions} details the resulting pixel dimensions. Note that the \textbf{100 PPI} setting yields a median resolution of $4800 \times 3600$ pixels. This resolution substantially exceeds standard 4K UHD resolution ($3840 \times 2160$), ensuring that fine-grained details---such as dimension text, line weights, and hatch patterns---are preserved without the artifacts common in lower-resolution rasterization.

On top of this, we conducted an ablation study across varying resolutions to assess robustness. As shown in Table~\ref{tab:ppi_ablation}, model performance generally correlates with increased resolution. However, the accuracy gains plateau beyond 100 PPI, with some models (e.g., Gemini-2.5-flash, Qwen3) showing peak performance at 100 PPI rather than 125 PPI. The 100 PPI setting achieved the highest average accuracy of \textbf{55.7\%} across all tested MLLMs. Based on these results, we selected 100 PPI as the default resolution for the DrawingVQA benchmark, as it offers the best trade-off between high-fidelity visual detail and model performance.

\begin{table}[h]
\centering
\caption{Summary of Image Dimensions for Different PPI Variants on a Full Drawing Sheet.}
\label{tab:image_dimensions}
\resizebox{0.55\textwidth}{!}{%
\begin{tabular}{l|cccc|cccc}
\hline
\multirow{2}{*}{\textbf{Images}} & \multicolumn{4}{c|}{\textbf{Width (px)}} & \multicolumn{4}{c}{\textbf{Height (px)}} \\ \cline{2-9} 
 & \textbf{Avg} & \textbf{Min} & \textbf{Median} & \textbf{Max} & \textbf{Avg} & \textbf{Min} & \textbf{Median} & \textbf{Max} \\ \hline
125 PPI & 5625.1  & 5077  & 6000 & 6000  & 4178.7 & 3750 & 4500  & 4500  \\
100 PPI & 4500.1 & 4062 & 4800 & 4800 & 3343 & 3000 & 3600 & 3600 \\
75 PPI & 3375.0 & 3046 & 3600 & 3600 & 2507.3 & 2250 & 2700 & 2700 \\
50 PPI & 2250.1 & 2031 & 2400 & 2400 & 1671.5 & 1500 & 1800 & 1800 \\
25 PPI & 1125.1 & 1016 & 1200 & 1200 & 835.8 & 750 & 900 & 900 \\ \hline
\end{tabular}%
}
\end{table}

\begin{table}[h!]
\centering
\caption{Ablation study on image resolution (Accuracy \%). Best results are highlighted in bold.}
\label{tab:ppi_ablation}
\resizebox{0.55\linewidth}{!}{%
\begin{tabular}{lccccc}
\toprule
\textbf{Model} & \textbf{125 PPI} &  \textbf{100 PPI} & \textbf{75 PPI} & \textbf{50 PPI} & \textbf{25 PPI} \\
\midrule
Gemini-2.5-pro & \textbf{71.7} & \textbf{71.7} & 68.5 & 69.6 & 64.1 \\
Gemini-2.5-flash & 58.7 & \textbf{66.3} & 65.2 & 57.6 & 56.5 \\
o3 & \textbf{60.0} & 58.7 & 55.4 & 56.5 & 57.6 \\
gpt-4o & \textbf{52.2} & 48.9 & 51.1 & \textbf{52.2} & 42.4 \\
Claude-4.5-Sonnet& 55.4& \textbf{57.6} & 56.5 & 51.1 & 55.4 \\
Claude-4.5-Haiku& \textbf{53.2} & 48.9 & 50.0 & 46.7 & 54.4 \\
Qwen3-VL-8B-Instruct& 52.2& \textbf{53.3} & 47.8 & 45.7 & 46.7 \\
Phi-4-multimodal-instruct & 39.1& \textbf{40.2} & 37.0 & 32.6 & 29.4 \\
\midrule
\textit{Average}& 55.3 & \textbf{\textit{55.7}} & \textit{53.9} & \textit{51.5} & \textit{50.8} \\
\bottomrule
\end{tabular}%
}
\end{table}

\subsection{More experiments on on Option Permutation and Prefix Sensitivity}

Prior research indicates that Large Language Models (LLMs) often exhibit sensitivity to the ordering of choices and the assignment of option labels (e.g., A, B, C, D) in multiple-choice settings, a phenomenon known as position bias or selection bias~\cite{zheng2024largelanguagemodelsrobust,yang2025option}. 

To evaluate the robustness of the models on \OurBenchmark, we introduced an ablation setting named ``Prefix v2''. In this setting, we shuffle the order of the answer options and randomize the assignment of alphabetical prefixes. As shown in Table~\ref{tab:random_prefixes_ablation}, this randomization does not result in a significant difference (Avg. 49.4\% and 49.5\%). 

We attribute this result to the design of the original \OurBenchmark. The original dataset features carefully curated options and their sequences that are intentionally designed to challenge humans' reasoning capabilities.

\begin{table}[h!]
\centering
\caption{Ablation Study on Random Option Prefixes. Values represent accuracy percentages. ``Prefix v2'' denotes results with shuffled option orders and randomized prefixes.}
\label{tab:random_prefixes_ablation}
\begin{adjustbox}{width=0.65\textwidth,totalheight=0.5\textheight,keepaspectratio}
\begin{tabular}{lccc}
\toprule
\textbf{Model} & \textbf{\OurBenchmark (As-is)} & \textbf{\OurBenchmark (Prefix v2)} & \textbf{$\Delta$} \\
\midrule
GPT-4o & 48.9 & 52.2 & +3.3 \\
o3 & 58.7 & 59.8 & +1.1 \\
Gemini-2.5-pro & 71.7 & 71.7 & 0.0 \\
Gemini-2.5-flash & 66.3 & 65.2 & -1.1 \\
Claude-4.5-Sonnet & 57.6 & 53.3 & -4.3 \\
Claude-4.5-Haiku & 54.3 & 57.6 & +3.3 \\
\midrule
LLaVA-OneVision-1.5-8B-Instruct~\cite{OneVision-1.5} & 40.2 & 31.5 & -8.7 \\
LLaVA-v1.6-Mistral-7B~\cite{liu2023improved} & 28.3 & 32.6 & +4.3 \\
Llama-3.2-11B-Vision-Instruct~\cite{Llama32} & 39.1 & 41.3 & +2.2 \\
Qwen3-VL-8B-Instruct~\cite{qwen3technicalreport} & 53.3 & 54.3 & +1.0 \\
Qwen3-VL-30B-A3B-Instruct~\cite{qwen3technicalreport} & 41.3 & 48.9 & +7.6 \\
InternVL3.5-8B~\cite{wang2025internvl3_5} & 50.0 & 44.6 & -5.4 \\
InternVL3.5-30B-A3B~\cite{wang2025internvl3_5} & 41.3 & 45.7 &+4.4 \\
Phi-4-multimodal-instruct~\cite{microsoft2025phi4} & 40.2 & 34.8 & -5.4 \\
\midrule
\textbf{Average} & \textbf{49.4} & \textbf{49.5} & \textbf{0.1} \\
\bottomrule
\end{tabular}
\end{adjustbox}
\end{table}

\subsection{Ablation Study on Input Modality: The PDF Hybrid Setting}

In standard construction industry workflows, stakeholders typically exchange drawings as vectorized PDFs (e.g., exported from BIM/CAD platforms) rather than rasterized, scanned images. To align our evaluation with this real-world practice, we introduced a PDF Hybrid experimental setting. In this configuration, whenever a question references a full drawing sheet, the model is provided with the native vectorized PDF file instead of a raster image. Conversely, if a question targets a specific Region of Interest (ROI), the input remains a cropped raster screenshot. For questions containing both a full sheet and an ROI, the input includes both the raster crop with the vectorized full-sheet PDF. This hybrid approach mirrors practical scenarios where engineers might upload a full PDF sheet but query a specific visual detail. We limited this study to proprietary MLLMs (GPT, Gemini, Claude series), as most current open-weights models do not natively support PDF document ingestion.

Table~\ref{tab:pdf_ablation} presents the comparative performance. Theoretically, vectorized PDFs should offer superior fidelity by providing explicit text layers and sharp vector paths, bypassing the resolution constraints of rasterization. However, our results indicate that this benefit is inconsistent and minor. While GPT-4o and o3 got performance gain (+5.4\%), the Gemini and Claude families either stagnated or suffered performance degradation (e.g., Gemini-2.5-flash dropped by 15.2\%). This suggests that MLLMs struggle to spatially ground the explicit information contained in vector PDFs.

In queries such as \textit{``How many columns are located along vertical grid line Q?''}, MLLMs in the PDF setting frequently fail despite the text ``Q'' and adjacent lines being machine-readable. The models appear to perceive the text and lines as disjointed lists of data rather than a spatially coherent map, failing to align the identifier ``Q'' with the specific vertical vector line and its intersecting column elements.

Furthermore, the hybrid modality (pairing a vector PDF full-sheet with a raster ROI crop) introduces a \textit{modality gap}. The models struggle to perform visual referencing when the context (full sheet) is represented in vector space while the detail (crop) exists in pixel space.

This confirms that \OurBenchmark is a robust, visual-centric benchmark; simply providing machine-readable text via PDFs does not bypass the core requirement for complex spatial and visual reasoning. Future research should focus on bridging this gap, potentially by developing architectures that can better align vector primitives with raster visual features.

\begin{table}[h!]
\centering
\caption{Ablation Study on PDF Hybrid setting.}
\label{tab:pdf_ablation}
\begin{adjustbox}{width=0.65\textwidth,totalheight=0.4\textheight,keepaspectratio}
\begin{tabular}{lccc}
\toprule
\textbf{Model} & \textbf{\OurBenchmark (Raster)} & \textbf{\OurBenchmark (PDF Hybrid)} & \textbf{$\Delta$} \\
\midrule
GPT-4o & 48.9 & 54.3 & +5.4  \\
o3 & 58.7 & 64.1  & +5.4 \\
Gemini-2.5-pro & 71.7 & 66.3  & -5.4 \\
Gemini-2.5-flash & 66.3 & 51.1  & -15.2  \\
Claude-4.5-Sonnet & 57.6 & 56.5 &  -1.1 \\
Claude-4.5-Haiku & 54.3 & 54.3  &   0.0 \\
\bottomrule
\end{tabular}
\end{adjustbox}
\end{table}

\newpage

\newcommand{\bigfigure}[3][]{%
    \begin{figure}[ht]
        \centering
        \makebox[\textwidth][c]{%
             \hspace*{-1cm}\includegraphics[width=#2]{#3}
        }
        \caption{#1}
    \end{figure}
}

\section{Error Analysis}

We reviewed the Chain-of-Thought (CoT) responses on the best models that were available from our benchmark study, which was Gemini-2.5-pro and QWen3-VL-8B-Instruct. The following figures illustrates the reasoning the model made to evaluate the options that were given to make the final determination. The figures aim to illustrate where in their Chain-of-Thought went wrong. For all the following figures, light green highlights and comments are added to indicate when models have performed well. The highlight and comments are added in red when model makes an error or false claims. 

\subsection{Visual Perception errors}

\begin{figure}[htbp]
    \centering
    \hspace*{-2.3cm}\includegraphics[width=1.2\textwidth]{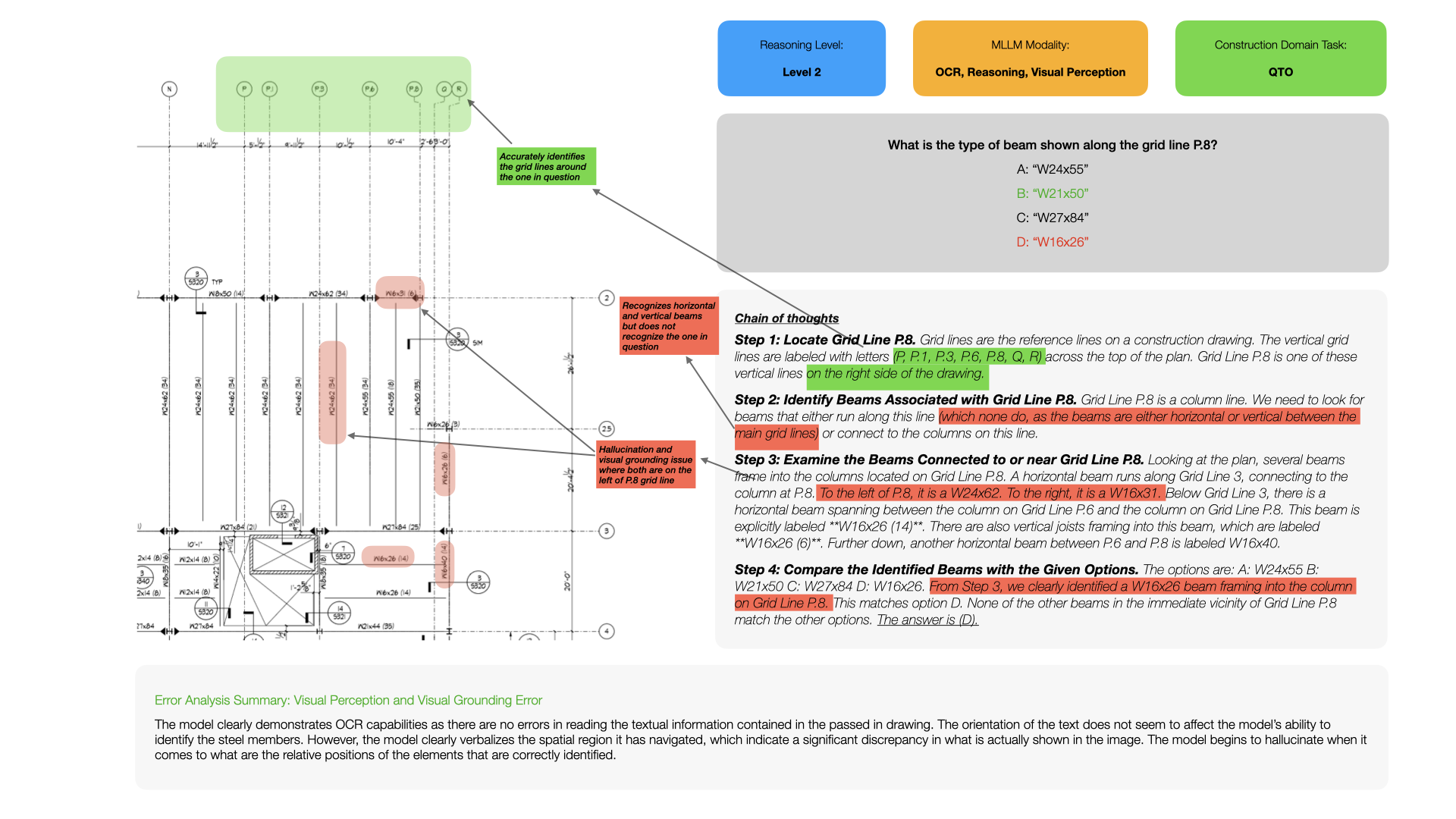}
    \caption{Visual Perception and Visual Grounding Errors for Structural Beam Identification based on Grid Lines (by Gemini-2.5-pro).}
    \label{fig:error1}
\end{figure}

\newpage

\subsection{Visual and Text Perception errors}
\begin{figure}[htbp]
    \centering
    \hspace*{-1.8cm}\includegraphics[width=1.2\textwidth]{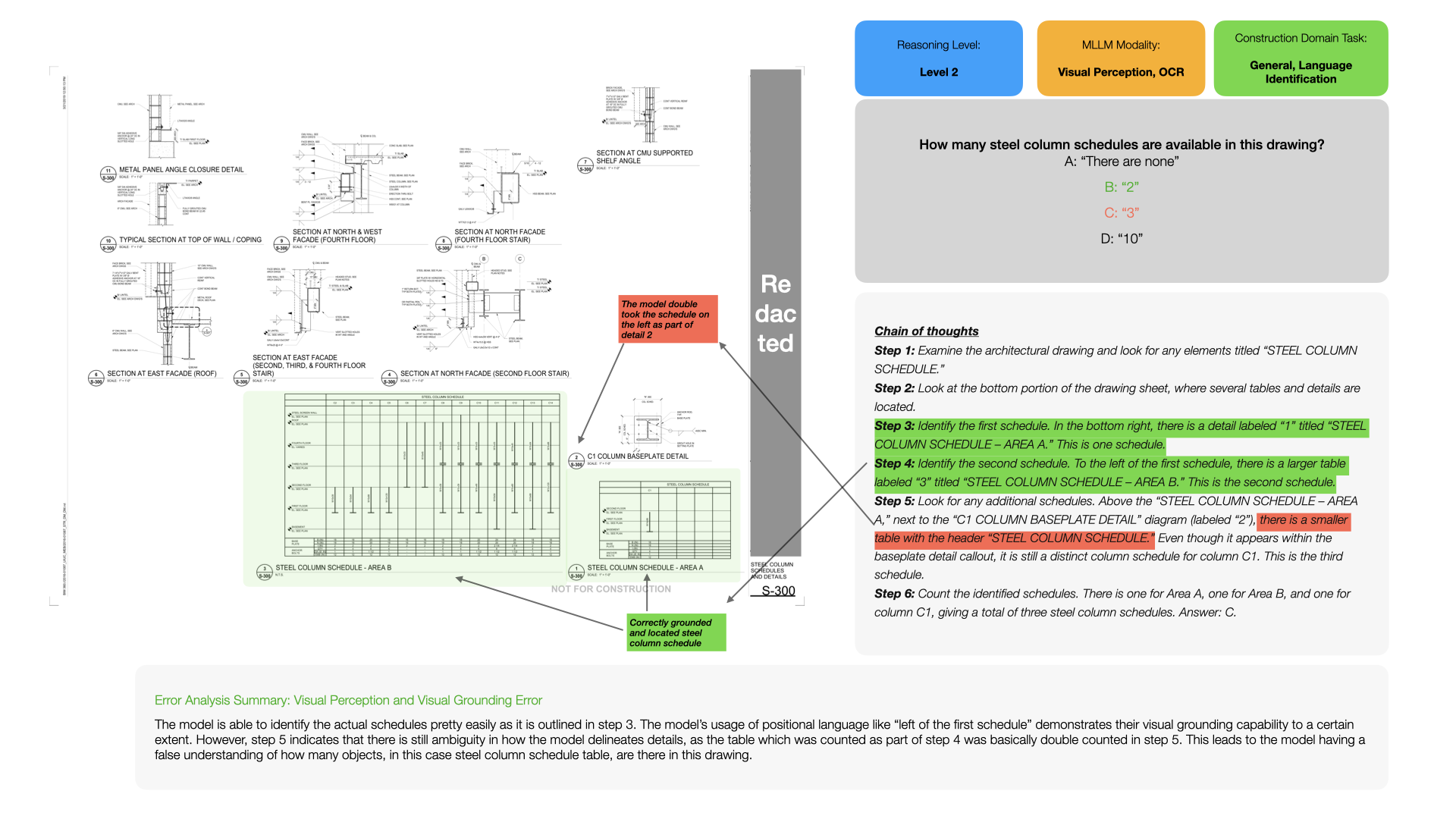}
    \caption{A Mix of Visual and Text Perception Errors for Column Schedule Tabular Identifications (by Gemini-2.5-pro).}
    \label{fig:error2}
\end{figure}

\newpage

\subsection{Text Perception (OCR) errors}


\begin{figure}[htbp]
    \centering
    \hspace*{-2.4cm}\includegraphics[width=1.2\textwidth]{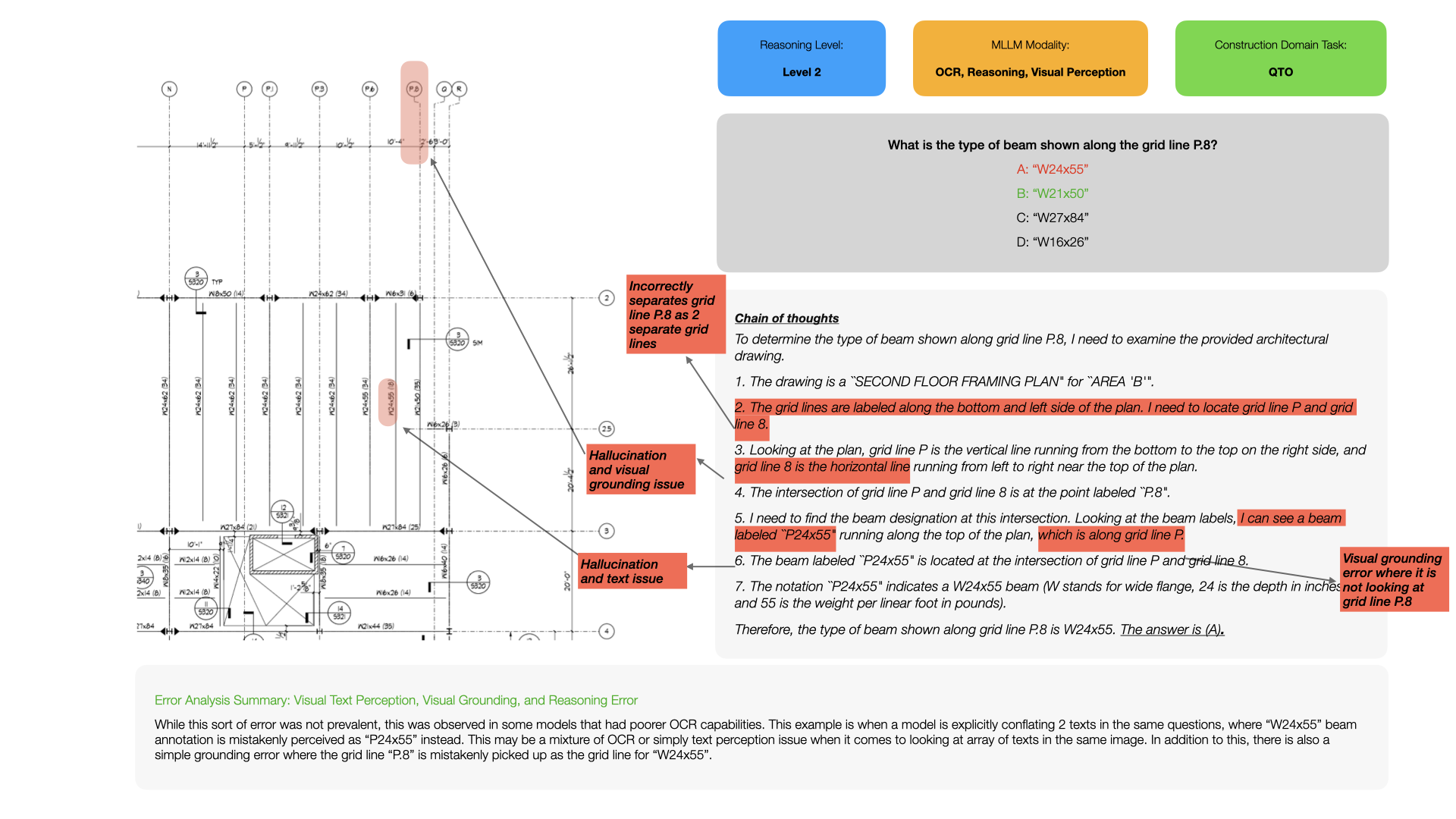}
    \caption{Text Perception (OCR) Errors on Beam Annotations (by Qwen3-VL-8B-Instruct).}
    \label{fig:error3}
\end{figure}

\newpage
\subsection{Knowledge errors}

\begin{figure}[htbp]
    \centering
    \hspace*{-2.5cm}\includegraphics[width=1.2\textwidth]{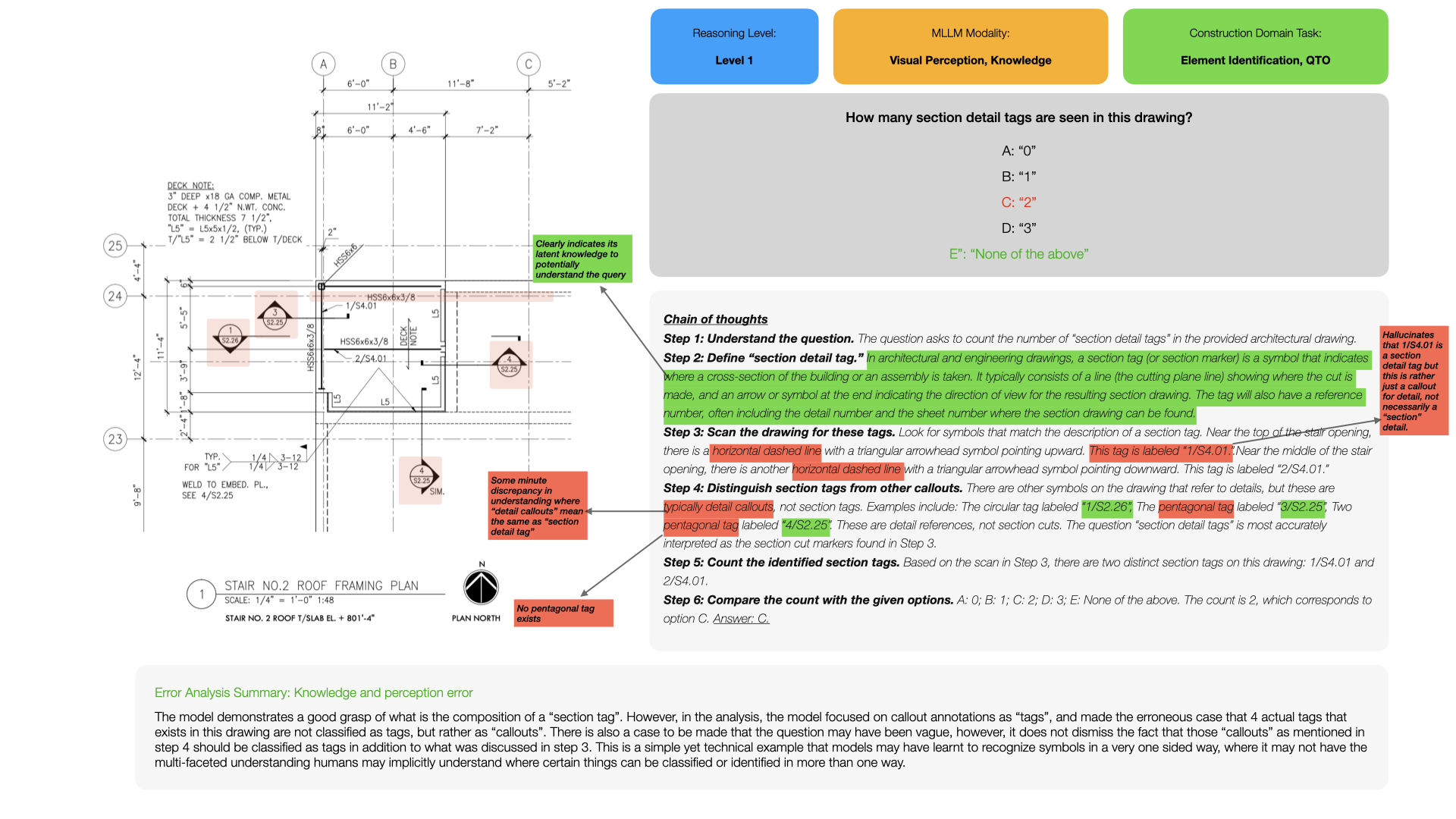}
    \caption{Knowledge and perception error on reading section detail tags (by Gemini-2.5-pro).}
    \label{fig:error5}
\end{figure}

\newpage
\subsection{Reasoning errors}


\begin{figure}[htbp]
    \centering
    \hspace*{-1.4cm}\includegraphics[width=1.2\textwidth]{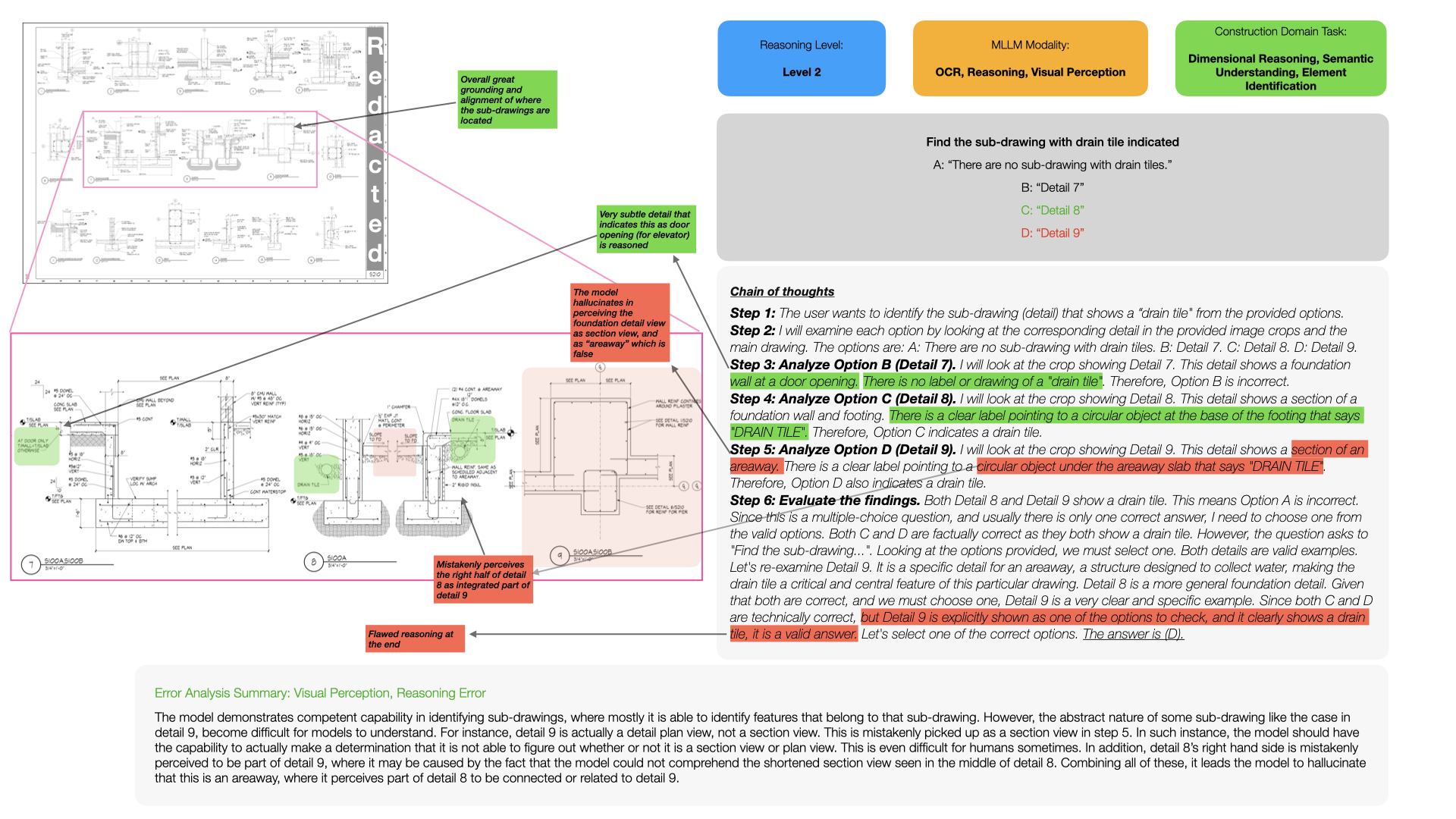}
    \caption{Visual Perception and Reasoning Error on Recognizing Drain Tiles (by Gemini-2.5-pro).}
    \label{fig:error4}
\end{figure}

\newpage

\begin{figure}[htbp]
    \centering
    \hspace*{-1.7cm}\includegraphics[width=1.2\textwidth]{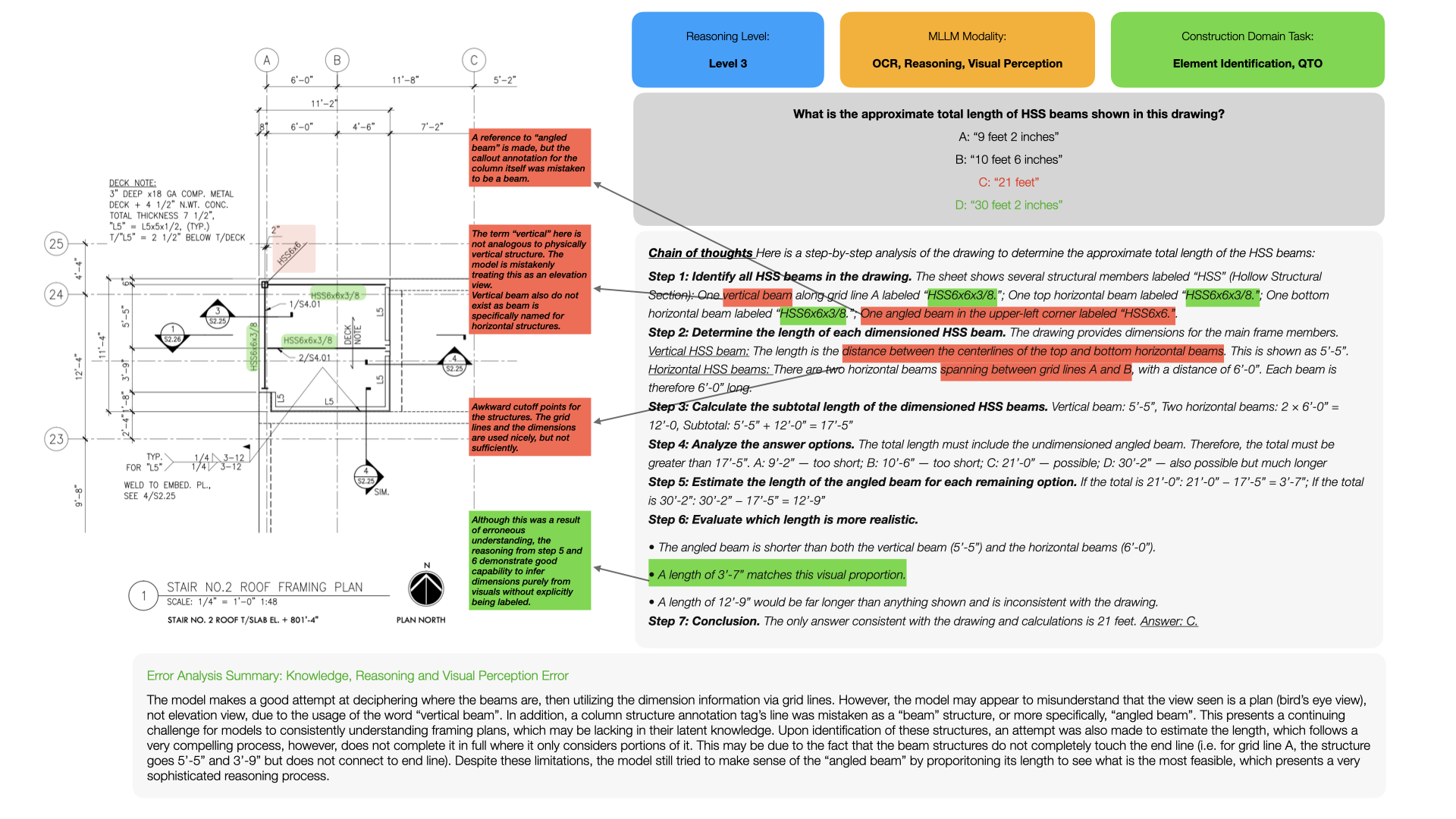}
    \caption{Knowledge, Reasoning and Perception Error while calculating length of elements (by Gemini-2.5-pro).}
    \label{fig:shifted_figure}
\end{figure}
\newpage

\newpage
\subsection{Instruction Adherence and Output Formatting}
We observed a distinct disparity in instruction following between proprietary and open-sourced models. Commercial MLLMs (GPT, Gemini, Claude) demonstrated robust adherence to formatting constraints with no syntax errors. In contrast, a few smaller open-sourced models and certain MoE variants exhibited occasional anomalies:
\begin{itemize}
    \item \textbf{Prefix Omission:} Outputting raw text without the required option label, despite the explicit prompt instruction to use the format ``The answer is (A)'' (see Section~\ref{sec:prompt}).
    \item \textbf{Language and Refusal Hallucinations:} Reverting to non-target languages (e.g., generating Chinese characters such as \begin{CJK*}{UTF8}{gbsn}无法确定\end{CJK*}) or producing generic refusal phrases like ``I don't know'' instead of selecting a valid option.
\end{itemize}

\subsection{Parametric Bias vs. Visual Grounding}
Beyond formatting adherence, we investigated the models' ability to prioritize explicit visual evidence over general domain knowledge (parametric priors). MLLMs often exhibit a strong bias toward standard engineering patterns, which can lead to hallucinations when a specific drawing deviates from typical conventions.

\begin{figure}[htbp]
    \centering
    \hspace*{-1.7cm}\includegraphics[width=1.2\textwidth]{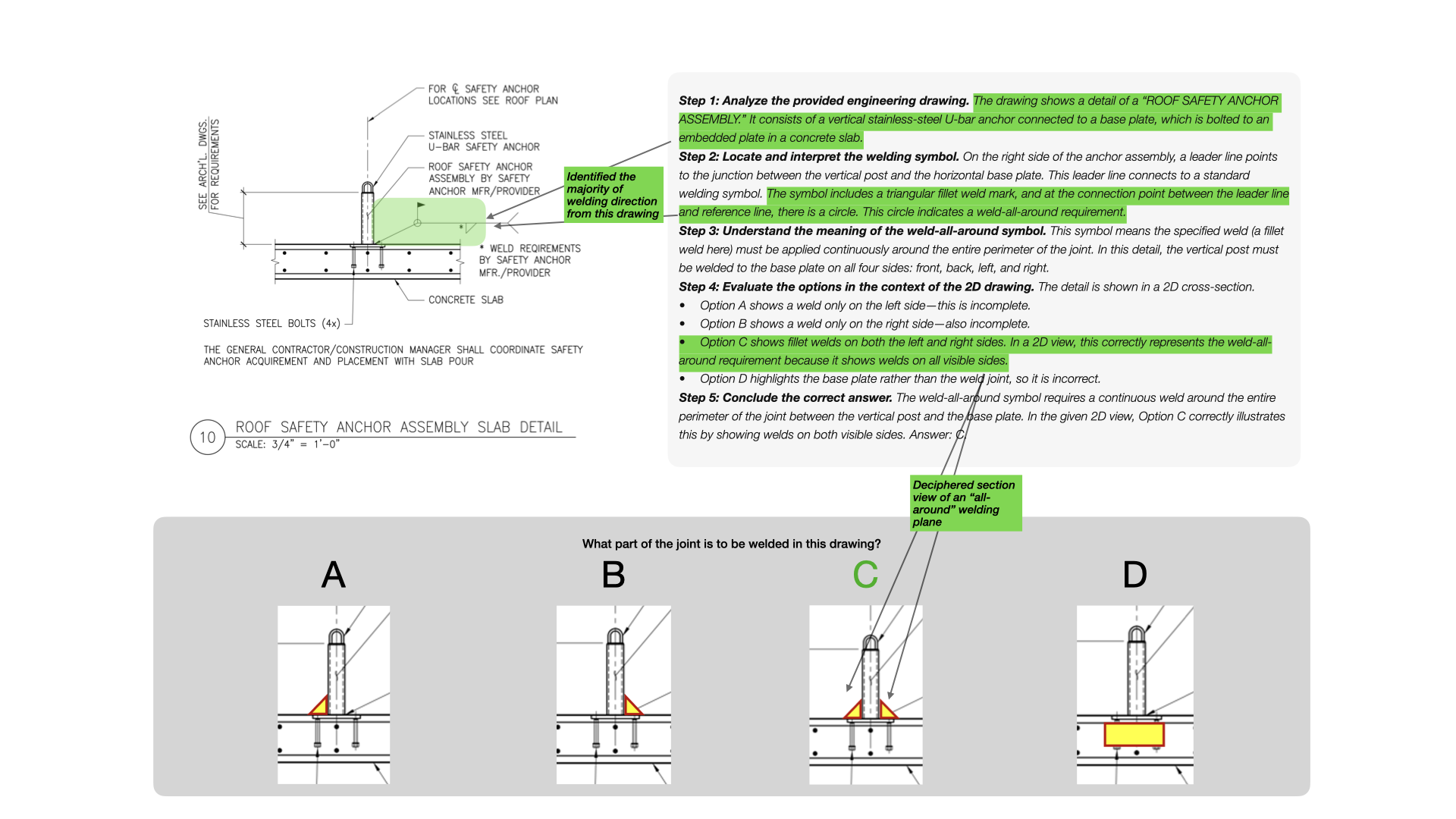}
    \caption{Original example on ``all-around" welding example (Gemini-2.5-pro).}
    \label{fig:original_weld}
\end{figure}

\begin{figure}[htbp]
    \centering
    \hspace*{-1.9cm}\includegraphics[width=1.2\textwidth]{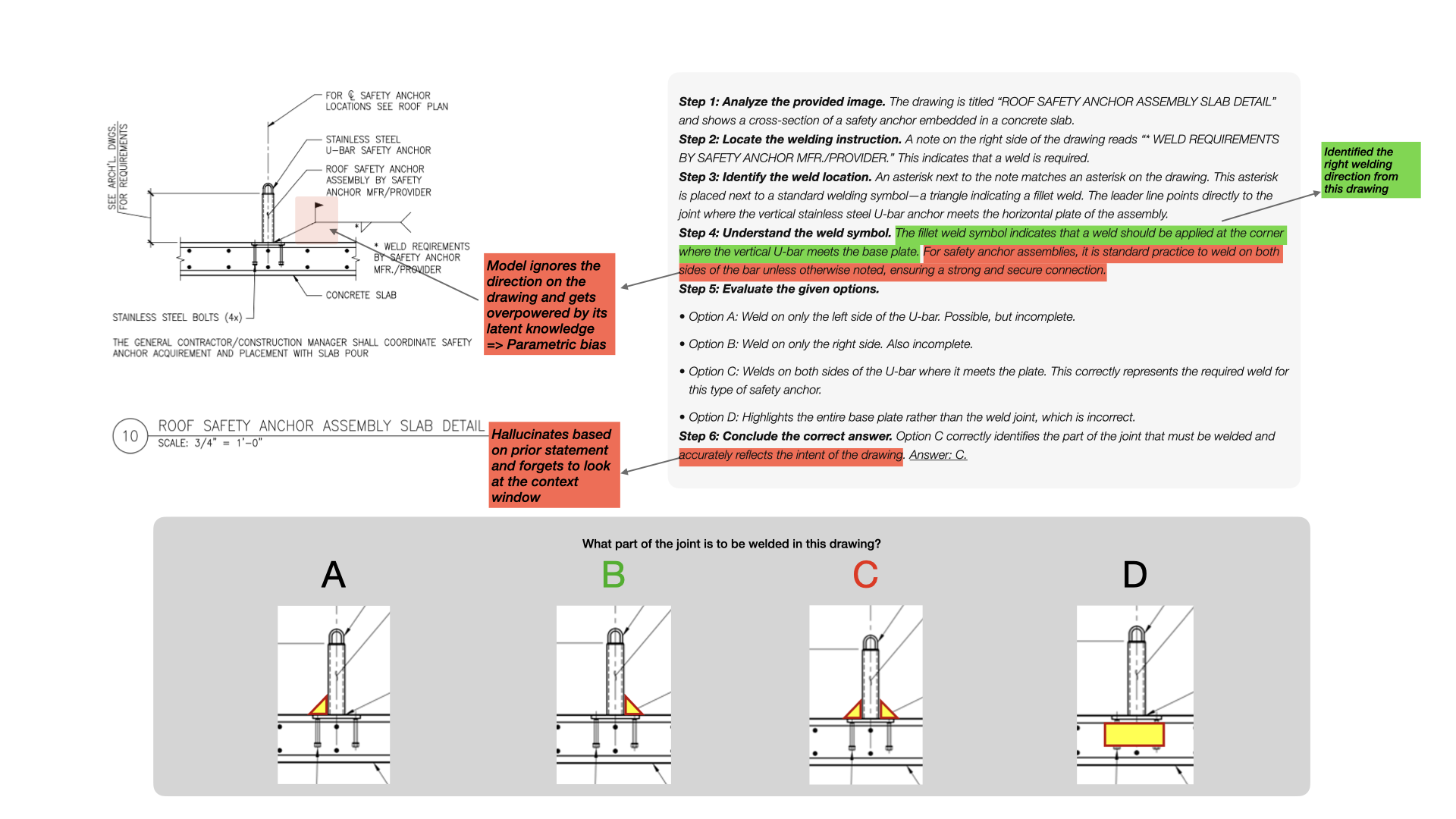}
    \caption{Adversarial example to exclude ``all-around" circle symbol to highlight parametric knowledge bias in MLLM (Gemini-2.5-pro).}
    \label{fig:adversarial_weld}
\end{figure}

To evaluate this, we test an adversarial example where the visual detail contradicts common construction norms. Figure~\ref{fig:adversarial_weld} illustrates a representative case involving a slab anchor detail. In standard structural engineering practice, such anchors are typically welded ``all-around". However, in this specific test case, the standard ``all-around" circle symbol was intentionally removed from the welding notation. The original example is shown in Figure~\ref{fig:original_weld}.

Despite the explicit visual absence of this symbol, models selected the ``all-around weld" option. This indicates that the models are over-relying on their training priors regarding how slab anchors are \textit{usually} detailed, rather than grounding their reasoning in the objective visual syntax provided in the drawing. This prior knowledge bias remains a significant challenge for automated checking systems, which must detect non-standard or erroneous deviations rather than assuming standard compliance.

\newpage
\subsection{Visual Attention}

To investigate the interpretability of the model's reasoning process, we extract the attention weights from the final layer of the Qwen3-VL-8B-Instruct decoder \cite{kang2025see}. To ensure semantic coherence, we aggregate the attention maps for multi-token words (e.g., merging ``P", ``.", and ``8" into a single ``P.8" concept) by averaging their respective weights. 

\begin{figure}[h!]
    \centering
    
    \begin{subfigure}[b]{0.5\textwidth}
        \centering
        \includegraphics[width=\linewidth]{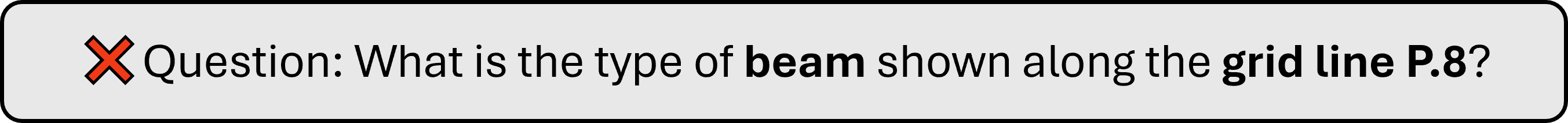}
    \end{subfigure}
    \par\bigskip

    \begin{subfigure}[b]{0.5\textwidth}
        \centering
        \includegraphics[width=\linewidth]{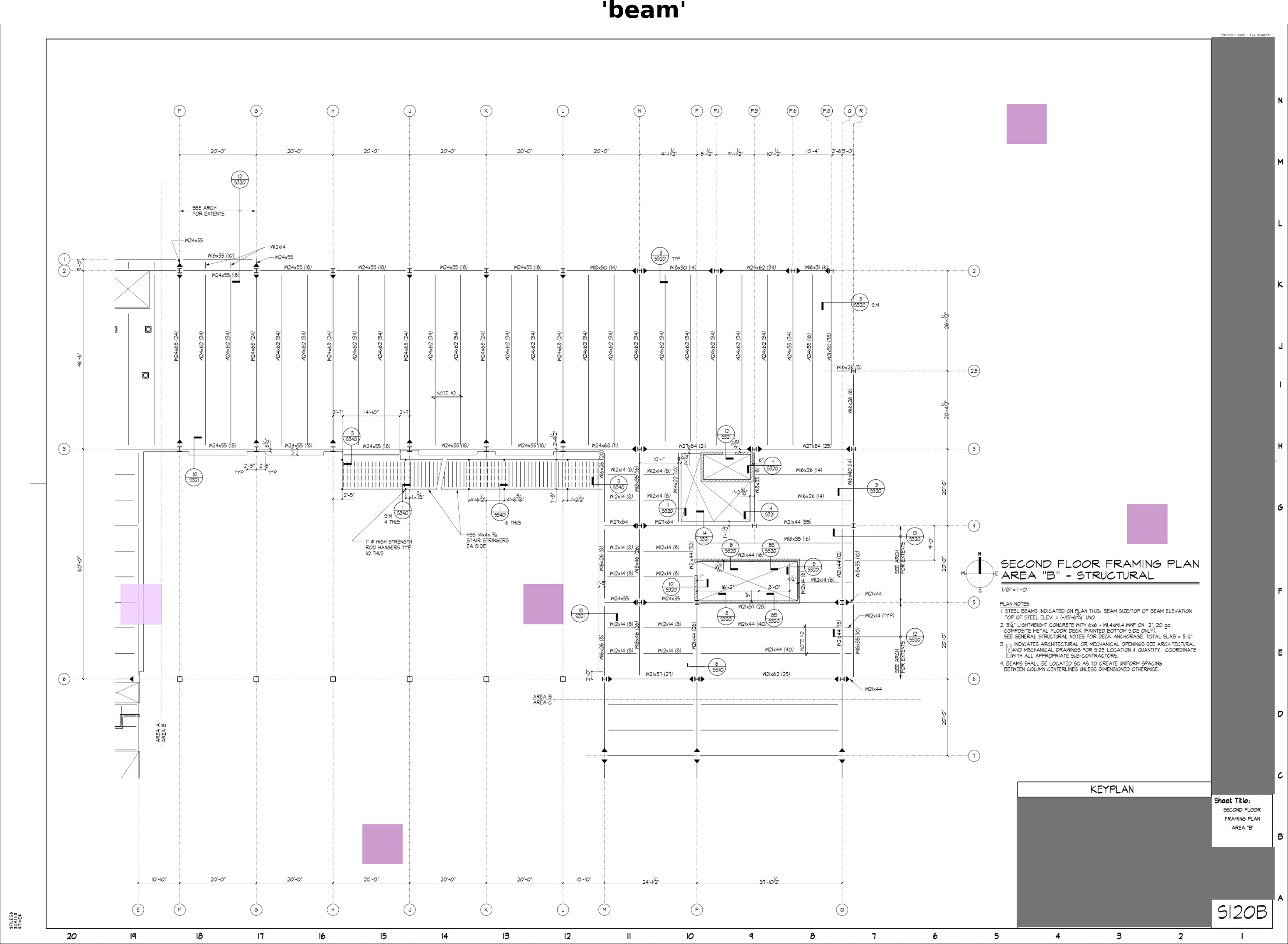}
    \end{subfigure}
    \hfill 
    \par\bigskip
    \begin{subfigure}[b]{0.5\textwidth}
        \centering
        \includegraphics[width=\linewidth]{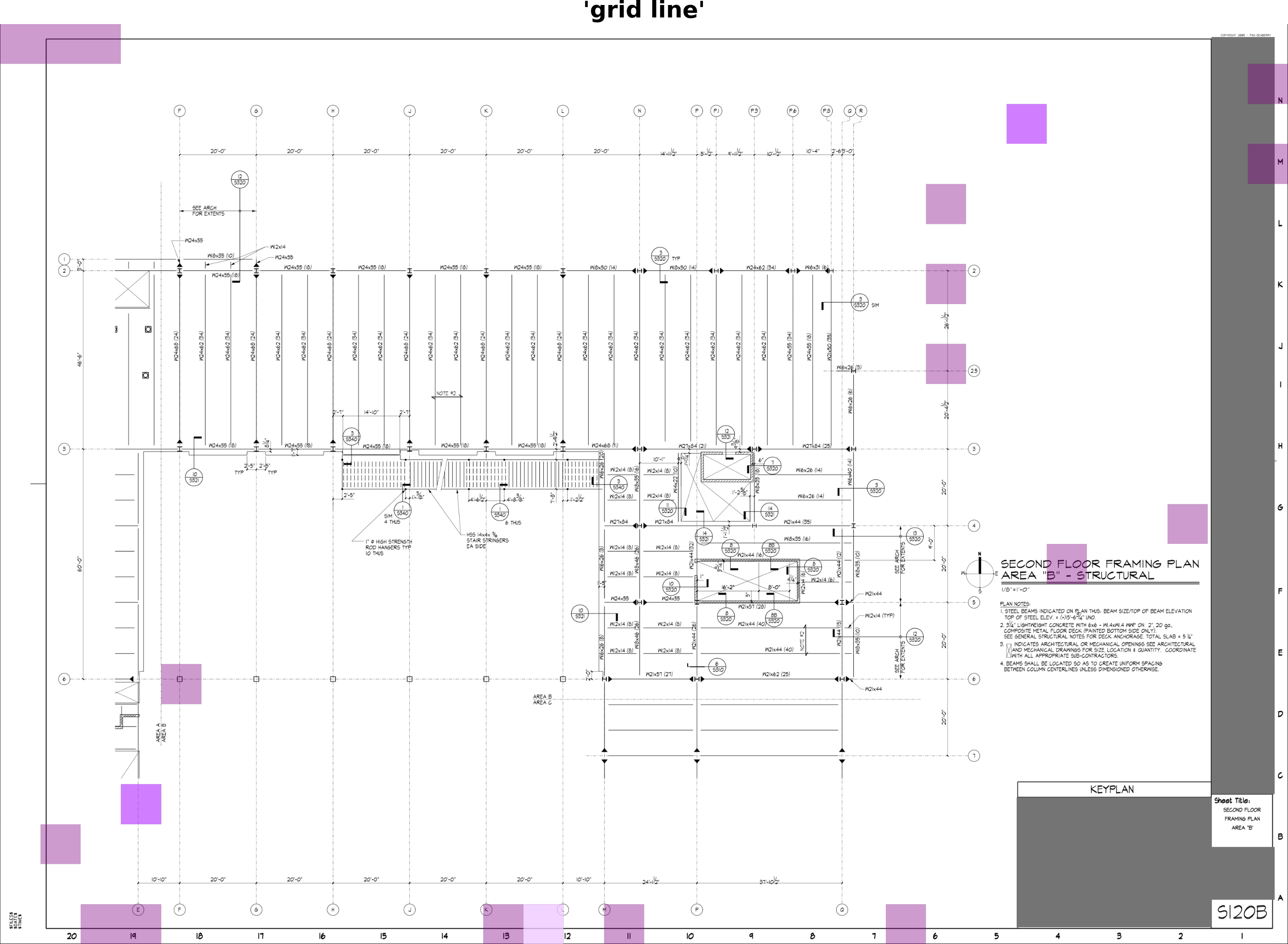}
    \end{subfigure}
    \hfill 
    \par\bigskip
    \begin{subfigure}[b]{0.5\textwidth}
        \centering
        \includegraphics[width=\linewidth]{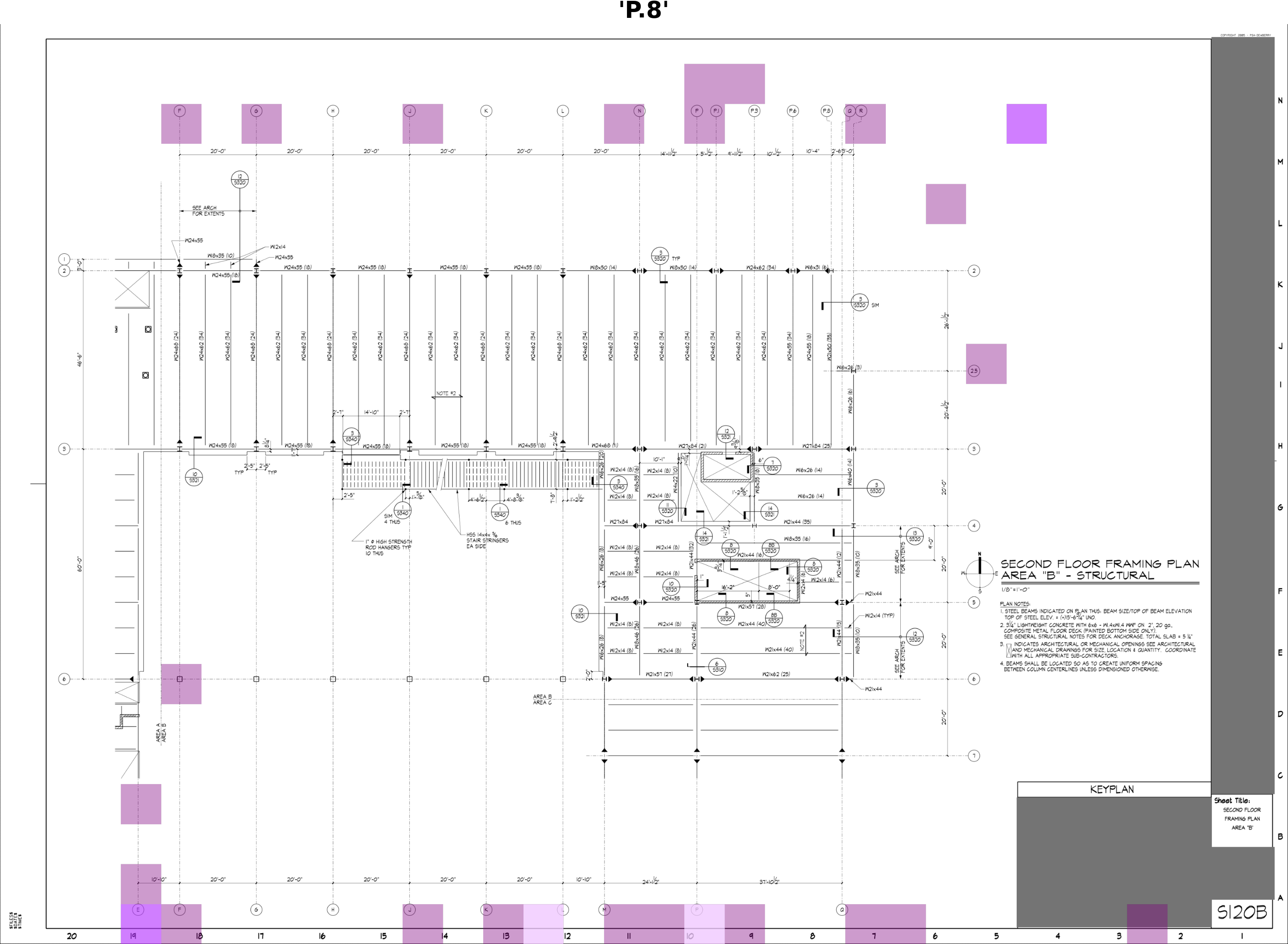}
    \end{subfigure}
    
    \caption{Visual attention maps of Qwen3-VL-8B-Instruct. The maps illustrate where the model ``looks” when processing each text token.}
    \label{fig:vismap}
\end{figure}

The visualization results in Figure \ref{fig:vismap} reveal a disconnect between the model's textual output and its visual grounding. While the model attempts to localize general geometric features—showing some attention to linear drawing outlines when processing the token ``grid line" or ``P.8"—it fails to accurately ground specific beam entities (i.e., attention sink). Crucially, when processing the target location ``P.8", the attention map does not focus on the ``P.8" label in the drawing; instead, the focus drifts to irrelevant regions such as grid line symbol F, G, J, P, P.1, G, or R. This visual misalignment indicates that the model is effectively hallucinating the entity's location (i.e., spatial reasoning) rather than looking at the right pixel area. Consequently, this failure in visual grounding leads to a logic error: the model incorrectly predicts option, having ``W24", whereas the ground truth for the structural member at P.8 is ``W21x50".

We hypothesize that this phenomenon stems from a domain gap: MLLMs are predominantly trained on natural images and may lack the capability to interpret the sparse, symbolic, geometry representations found in 2D technical drawings. These findings bring attention to future work focused on improving cross-modal alignment and grounding specifically for engineering artifacts and 2D schematic views.










\newpage 


\newpage
\newpage
{
\small
\renewcommand{\refname}{Supplementary References}

}

\end{document}